\documentclass[10pt,twocolumn,letterpaper]{article}

\usepackage{wacv}
\usepackage{times}
\usepackage{epsfig}
\usepackage{graphicx}
\usepackage{amsmath}
\usepackage{amssymb}
\usepackage{booktabs}
\usepackage[accsupp]{axessibility}

\def\wacvPaperID{1814} 

\wacvapplicationstrack 

\wacvfinalcopy 

\usepackage{hyperref}

\begin{document}

\title{Image-Text Pre-Training for Logo Recognition}

\author{Mark Hubenthal\\ {\tt\small mhubenth@amazon.com}\\ Amazon Inc. \and Suren Kumar\\ {\tt\small ssurkum@amazon.com}\\
Amazon Inc.
}

\maketitle

\thispagestyle{empty}

\begin{abstract}
Open-set logo recognition is commonly solved by first detecting possible logo regions and then matching the detected parts against an ever-evolving dataset of cropped logo images.
The matching model, a metric learning problem, is especially challenging for logo recognition due to the mixture of text and symbols in logos.
We propose two novel contributions to improve the matching model's performance: (a) using image-text paired samples for pre-training, and (b) an improved metric learning loss function.
A standard paradigm of fine-tuning ImageNet pre-trained models fails to discover the text sensitivity necessary to solve the matching problem effectively.
This work demonstrates the importance of pre-training on image-text pairs, which significantly improves the performance of a visual embedder trained for the logo retrieval task, especially for more text-dominant classes.
We construct a composite public logo dataset combining LogoDet3K, OpenLogo, and FlickrLogos-47 deemed OpenLogoDet3K47.
We show that the same vision backbone pre-trained on image-text data, when fine-tuned on OpenLogoDet3K47, achieves $98.6\%$ recall@1, significantly improving performance over pre-training on Imagenet1K ($97.6\%$).
We generalize the ProxyNCA++ loss function to propose ProxyNCAHN++ which incorporates class-specific hard negative images.
The proposed method sets new state-of-the-art on five public logo datasets considered, with a $3.5\%$ zero-shot recall@1 improvement on LogoDet3K test, $4\%$ on OpenLogo, $6.5\%$ on FlickrLogos-47, $6.2\%$ on Logos In The Wild, and $0.6\%$ on BelgaLogo.
\end{abstract}

\section{Introduction}
Logo Recognition, the problem of identifying both logo regions and their associated logo classes, has several essential application areas, including brand recognition, contextual advertising, and trademark infringement detection \cite{tursun2015metu}.  
Research in this field has closely tracked advances in general image recognition, given the similarities between the two problems.
The field has moved from hand-designed geometric invariants \cite{doermann} to Scale-Invariant Feature Transforms (SIFT) \cite{lowe1999object} and is now driven entirely by large deep neural networks \cite{Logos32Plus2017}.
Two unique factors separate logo recognition from the deep learning-based image recognition trajectory.
First, unlike typical object detection or image classification problems, new logo classes are constantly emerging as new companies are registered or contemporary styles are adopted.
Second, an archetypal logo class can have an abstract definition with high intra-class variation, including different colors, styles, texts, and image backgrounds.
Several research teams have addressed the first challenge by extracting a logo-agnostic bounding box and then leveraging metric learning to match detected regions against a potentially evolving set of logo images.
However, metric learning approaches have struggled to perform robustly on the logo region matching problem due to high intra-class variations and the text-heavy nature of logo regions \cite{Li2022WACV}.

We hypothesize that the lack of text sensitivity in deep neural networks for metric learning is due to constraints in the fine-tuning process. 
The success of deep learning models on computer vision problems is driven by pre-training on large open-source datasets such as ImageNet and then adapting network weights to a downstream task.
The performance of the resulting networks is heavily dependent on the characteristics of the pre-training dataset and the task similarity between fine-tuning and pre-training stages.
However, popular pre-training datasets such as ImageNet or MS-COCO have limited text present in the images they contain.
Thus, the current fine-tuning paradigm has severely limited performance on tasks that require text understanding, such as logo recognition and text-based visual question answering.
Researchers have attempted to overcome such challenges around text representation by explicitly extracting optical character recognition (OCR) information and representing it for image captioning and visual question answering \cite{yang2021tap,kant2020spatially,zhu2020simple}.
Text-aware embeddings are crucial for open-set logo recognition because logos often contain text content.
Li et al. \cite{Li2022WACV} proposed explicitly leveraging the brand name as a target to build a parallel text-in-image representation to bridge the gap between the representative capability of traditional vision-backbones and the demands of logo recognition.
Hu. et al. \cite{hu2020multimodal} proposed fusing explicit OCR with a textual description of an image in a multimodal fusion framework.
However, such approaches increase overall embedding size, require additional downstream supervision data, and achieve limited performance gains.

We propose building on the recent success of large multimodal pre-training to improve text sensitivity of metric learning network backbones.
Our solution, similar to existing works such as CLIP \cite{Clip2021} and ALIGN \cite{jia2021scaling}, leverages normalized softmax-based constrastive learning to pre-train vision networks that generate text-sensitive embeddings.
When adapted to the logo recognition problem, such networks significantly improve performance without requiring additional supervision, a separate text processing backbone, or OCR extraction.
We also improve the fine-tuning process of the metric learning backbone with a new, improved loss function, deemed ProxyNCAHN++, that can leverage hard negatives in or out of domain.
This helps to address the issue of very similar logos or the potential similarity between a logo and some out-of-domain designs or patterns.

In summary, the key contributions of our work are 
\begin{enumerate}
\item We propose using image-text multimodal pre-training to improve text sensitivity of metric learning backbones for open-set logo recognition.
\item We present a new loss function to increase inter-class distance despite significant intra-class variation. An ablation study is documented in Section \ref{subsec:ablation}.
\item We perform a rigorous evaluation on several open-source logo recognition datasets in Section \ref{subsec:publiccomparison}, where our solution surpasses state-of-the-art.
\end{enumerate}

\begin{figure*}[!ht]
  \centering
  \includegraphics[width=0.9\linewidth]{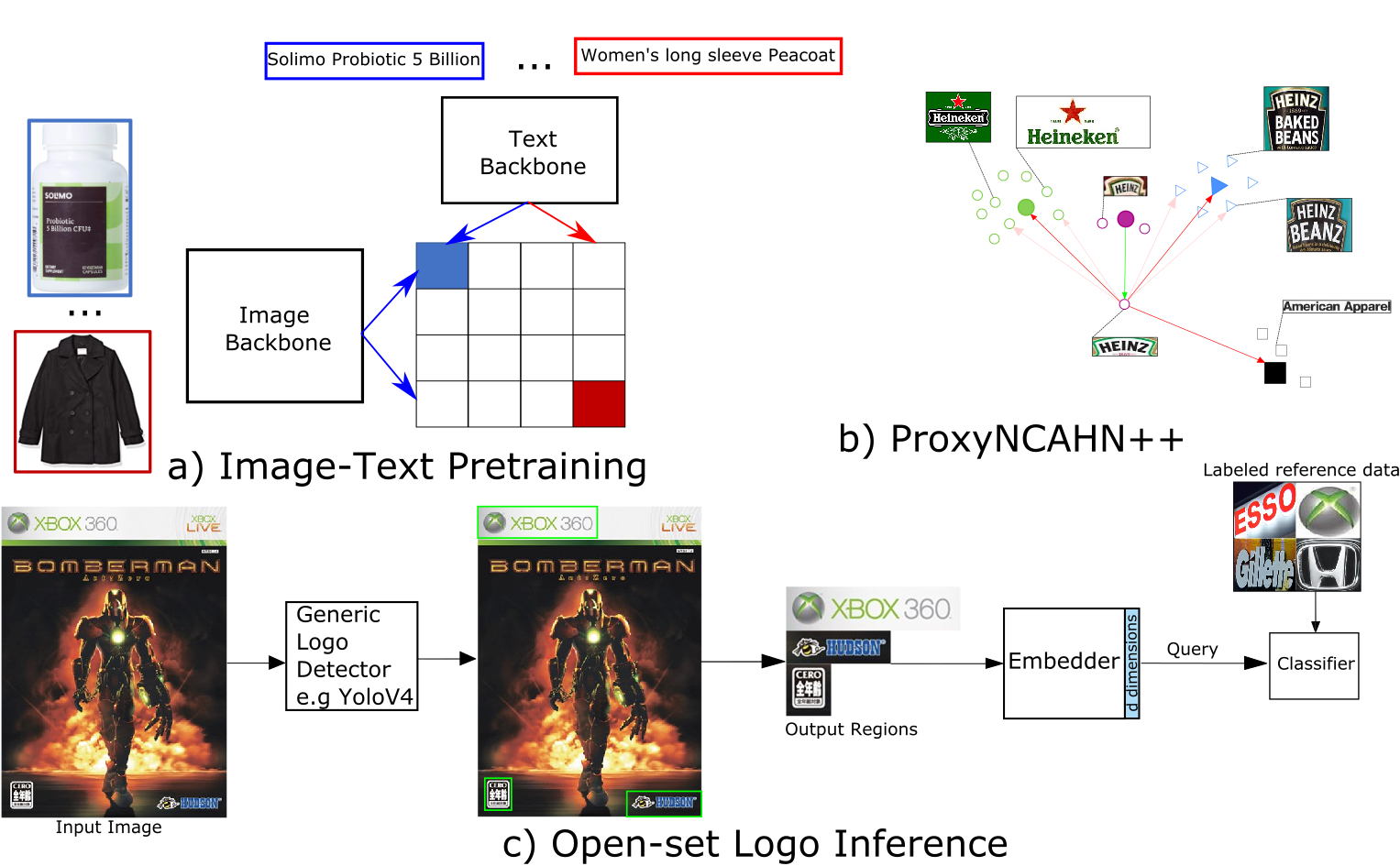}\label{fig:overall_system}
  \caption{Overall setup of the proposed open-set logo recognition pipeline. 
  a) Image-Text Pretraining: A batch of images with corresponding text titles is independently passed through a vision and a text backbone to extract embeddings. Both the backbones are trained by maximizing the alignment between the corresponding pairs ensuring a diagonally dominant dot product similarity matrix.
  b) The pre-trained image backbone is further fine-tuned to serve as an embedder for the logo recognition problem using a new metric loss function, ProxyNCAHN++. Filled-in shapes represent learned proxy vectors and unfilled shaped represent image embedding vectors.
  For an input image from class \textsf{h\_j\_heinz}, it is attracted to its own class' proxy (green arrow), pushed away from other class proxies (red arrows), and also pushed away from individual image embeddings belonging to assigned hard negative classes \textsf{heinz\_baked\_beans} and \textsf{heineken\_text} (pink arrows).
  c) Inference Pipeline: A generic logo detector extract all possible logo bounding boxes during inference. We then extract embeddings of these individual boxes (with the embedder trained in Step (b)) and compare against a dataset of evolving logos to compute the class labels.
  }
\end{figure*}

\section{Related Work\label{sec:relatedwork}}
\subsection{Logo Recognition}
The problem of logo recognition has closely tracked the progress in the broader image recognition field. 
In one of the early works of 1993, Doermann et al. \cite{doermann} designed many geometrical invariants such as image moments to match the features of a new logo image against a database. 
The transition to better hand-crafted features such as Scale-Invariant Feature Transforms (SIFT) \cite{lowe1999object} were readily adopted by the logo recognition community \cite{BelgaLogos09}.
Early attempts in the late 90s at leveraging neural networks \cite{Cesarini} for logo recognition problems achieved limited success.
That has changed in the past few years due to better neural network architectures \cite{ResNet2016}, training techniques \cite{kingma2014adam} and availability of large datasets \cite{deng2009imagenet} and GPUs.
Bianco et al. \cite{Logos32Plus2017} proposed building on the success of the Fast Region-Based Convolutional Network \cite{girshick2015fast} recognition pipeline by first sampling image regions and then classifying them into a logo class.
Su et. al. \cite{OpenLogo2018} utilized improvements in the object detection literature, specifically Faster R-CNN \cite{ren2015faster} and YOLO \cite{redmon2017yolo9000}.
They further relaxed the requirement of localized logo bounding boxes by synthetically compositing logo images onto background images.
Wang et al. \cite{LogoDet3K2022} introduced a large dataset of 3K logo categories and made improvements to the YOLO framework. 
These methods made substantial gains in logo-recognition performance but were only effective in the closed-set setting.
Adding any new logo at test time would require re-training such models.

Tuzko et al. \cite{tuzko2017open} proposed one of the first deep-learning-based open-set logo recognition methods. 
They use a Faster R-CNN \cite{ren2015faster} network to detect any logo region in the image and use an ImageNet-pre-trained network to match the logo region against a dataset of cropped logos.
Fehérvári et al. \cite{Istvan2019} proposed taking advantage of metric learning approaches to improve the matching process further.
This two-step paradigm of class-agnostic logo region localization and subsequent feature matching against an ever-evolving set of logos has become the norm \cite{Bastan2022}.
The key challenge is to ensure the robustness of the metric learning backbone, specifically its text sensitivity.
Typical text-invariance is due to initialization from ImageNet pre-trained networks and a lack of direct OCR supervision.
Some fixes have been proposed to close this gap. 
Li et al. \cite{Li2022WACV} proposed a separate branch of text feature extraction to generate the text contained in the logo. 
Hu. et al. \cite{hu2020multimodal} explicitly extract OCR information from the image and use an associated image description process through BERT \cite{devlin2018bert} to extract brand information in the image.
In contrast, we propose a simple change to the matching network's pre-training strategy, which does not require additional training data or increase in computational cost.
Our method uses image-text multimodal pre-training, which imbues the vision network backbone with OCR capabilities. 

\subsection{Image-Text Pretraining}
The use of attention-based methods in computer vision such as Vision Transformers \cite{dosovitskiy2020image} has led to many works that jointly model image and text.
Vision and language transformer (ViLT) \cite{kim2021vilt} proposed concatenating words with image patches to obtain a joint representation. 
ViLT achieved impressive results on multiple downstream applications, including visual question answering and image captioning.
Li et al. \cite{li2021align} demonstrated multimodal training on a noisy set of image-text correspondences without the need for detailed annotations.
We are particularly enthused by Contrastive Language-Image Pre-Training (CLIP) \cite{Clip2021} where the authors proposed a simple yet effective method for multimodal training.
CLIP constructs a large image-text dataset, WebImageText (WIT), by extracting images and their corresponding descriptions (such as alt text) and uses the resulting pairs to contrastively train image and text encoder backbones.
The image backbone thus trained beats all other pre-trained methods on datasets that require an understanding of the text in the image, such as HatefulMemes \cite{kiela2020hateful}.

In addition to experimenting with OpenAI's CLIP-trained vision transformer (ViT) model, we train an image-text backbone on a large dataset of product image-title pairs from an e-commerce website for comparison.
The trained network demonstrates significant improvement on logos that contain text, as shown by Figure \ref{fig:qualcomp}.
Next, we discuss the loss function used to fine-tune the matching network's weights.  

\subsection{Metric Learning Loss Function}
The three key attributes of any embedder/matcher are the network backbone and its pre-training, the loss function, and the metric used to compare the extracted embeddings.
Training a matching model aims to push the embeddings of related logo images closer while moving apart image embeddings of different logos.
Some early works such as pair contrastive loss \cite{hadsell2006dimensionality} were adopted in designing losses like the triplet loss \cite{tripletloss} to better structure the embedding space.
However, a common difficulty in pair-wise, triplet, or higher-pair loss functions has been the issue of sampling informative pairs in a batch.
ProxyNCA \cite{ProxyNCA2017} proposes an alternative way of modeling the class distribution, where a proxy represents a class, and images in a batch are pushed closer to their corresponding class proxies.
Teh et al. \cite{Teh2020} built several improvements into the original ProxyNCA method such as fast-moving proxies.
In this paper, we further improve on ProxyNCA++ by better capturing the boundaries of the class distribution.
Our loss function, ProxyNCAHN++, explicitly samples a set of hard negative images per class in each batch so as to better separate each class in feature space.

\section{System Design\label{sec:systemdesign}}
Our overall system follows the general paradigm in open-set logo recognition.
First, a general logo-region detector such as \cite{YoloV42020,redmon2017yolo9000} identifies axis-aligned bounding boxes likely to contain a logo region.
Then, a feature embedder module extracts a representation of each bounding box and matches it against a stored image representation of possible logo classes.
The two core contributions of this work are in pre-training the feature embedder module to imbue text sensitivity and in using a better loss function for training the feature embedder model.
Figure \ref{fig:overall_system} shows the high-level system design.
Next, we describe the two unique components of our design.

\subsection{OCR Sensitive Embedding}
We build on a simple yet effective design from CLIP \cite{Clip2021} to perform image-text pre-training.
A batch of $N$ image-text pairs are fed to separate vision and language processing modules.
The two modalities are trained by computing an $N\times N$ matrix of dot products between all possible image-text pairs.
Finally, supervision is applied by encouraging the matrix to be diagonally dominant, i.e., ensuring that embeddings of an input pair are closely aligned compared to all other pairings.
The sum of cross-entropy losses for each image-text pair generates the training signal for both vision and language backbones.
The resulting model, especially the vision backbone, is found to possess strong OCR discovery capabilities.
In particular, Appendix E.2 of \cite{Clip2021} documents $80.5\%$ accuracy of the visual encoder on the rendered SST-2 NLP dataset.
OCR discovery comes from having rich descriptions that reference scene text contained in an image, e.g. a caption with a brand name corresponding to an image containing that brand's logo.
Another advantage of this paradigm is the clear separation between the two modalities, which enables the evaluation of each modality independently.

The WIT dataset collected and used to train the CLIP model is not available publicly, but we can still use the model weights for fine-tuning.
We also use our own set of product image-title pairs scraped from an e-commerce website as a comparable image-text pre-training set.
Our experiments prove that even with a much smaller dataset than WIT used for pre-training, i.e., approximately 20 million product image-title pairs, we obtain an embedder that significantly outperforms one pre-trained with ImageNet data.

\subsection{Loss Function}
This section describes the primary loss function used and how we can incorporate complex negative samples explicitly.
\subsubsection{ProxyNCA++ Loss}
One of the recent standard approaches to training deep metric learning models is the Proxy NCA++ loss \cite{Teh2020}, which improves on ProxyNCA \cite{ProxyNCA2017}.
The similar normalized softmax loss proposed in \cite{zhai2019classification} also utilizes proxies.
The main idea is to use trainable proxy vectors to represent the ideal embedding of a given class (generalizable to multiple proxies per class in the dynamic assignment case).
As such, we no longer need to sample pairs or triplets of data points within a batch as is the case of a standard contrastive or triplet loss. 
We can train the model as a classification problem by looking at the distance from each image embedding to its corresponding proxy vector.

Let $\mathcal{P}$ be the set of all learned proxy vectors in $\mathbb{R}^{n}$.
For a given batch of images consider a positive class input image $x_{i} \in \mathbb{R}^{3 \times p_{y} \times p_{y}}$ with logo class $y_{i} \geq 0$, where $p_{x}, p_{y} \in \mathbb{N}$ are the input image width and height, respectively.
Further let $z_{i} \in \mathcal{P}$ denote the proxy vector for $x_{i}$, i.e. $z_{i}$ is closest to $f_{\theta}(x_{i})$ among all proxies with class $y_{i}$, where $f_{\theta}:\mathbf{R}^{3\times p_{y} \times p_{x}} \to \mathbf{R}^{n}$ is the embedding function with learned parameters $\theta$.
For $z_{1},z_{2} \in \mathbb{R}^{n}$ define
\begin{equation}
\label{eq:distance}
d(z_{1}, z_{2}) = \left\| \frac{z_{1}}{\|z_{1}\|_{2}} - \frac{z_{2}}{\|z_{2}\|_{2}} \right\|_{2}^{2},
\end{equation}
and let $g(z_{1}, z_{2}) = \exp\left(-\frac{d(z_{1}, z_{2})}{\sigma}\right)$, where $\sigma > 0$ is a temperature-scaling parameter.
The standard ProxyNCA++ loss for input $x_{i}$ is then
\begin{align}
L_{\mathrm{ProxyNCA}^{++}} & = -\log(P_{i})\\
P_{i} & = \frac{ g(f_{\theta}(x_{i}), z_{i}) }{\sum_{z \in \mathcal{P}} g(f_{\theta}(x_{i}), z) }. 
\label{eq:proxynca++}
\end{align}
Note the squared exponent in (\ref{eq:distance}) just as in \cite{Teh2020,ProxyNCA2017}, which we also observed performs better than the standard $L^{2}$ distance.

\subsubsection{Handling Hard Negatives}
In attempt to improve performance further, we consider emphasizing difficult negatives during training that belong to another positive class or no class at all.
We employ a batch sampling strategy where for each class $y_{i}$, a randomly selected difficult negative class $y_{j}$ is also included.
Then in the denominator of $P_{i}$ from (\ref{eq:proxynca++}), we can add the exponential distances from the embedding $f_{\theta}(x_{i})$ to all embeddings for class $y_{j}$ in the batch.
To make this more precise, we focus on the closed-set regime where test classes are the same as those in the training set.
Let $\mathcal{L} = \{y_{1}, \ldots, y_{k}\}$ be the set of all distinct classes, and let $\widetilde{\mathcal{L}} = \{\widetilde{y}_{1}, \ldots, \widetilde{y}_{k}\}$ be the set of all (background) negative class labels.
Assume we have a mapping $h:\mathcal{L} \ni y_{i} \mapsto \{h(y_{i})_{1}, \ldots, h({y}_{i})_{s_{i}}\} \subset \mathcal{L} \cup \widetilde{\mathcal{L}}$ that assigns each label $y_{i}$ a set of difficult hard negative classes (possibly empty).
Now define for each input image $x_{i}$ with positive class $y_{i} \in \mathcal{L}$ in a batch $\mathcal{B}$ the ProxyNCAHN++ loss:
\begin{equation}
L_{\mathrm{ProxyNCAHN}^{++}} = -\log(P_{i}^{HN})
\end{equation}
where
\begin{equation}
\label{eq:proxyncahn++}
P_{i}^{HN} \propto \frac{ g(f_{\theta}(x_{i}), z_{i})}{\begin{matrix}\sum_{z \in \mathcal{P}} g(f_{\theta}(x_{i}), z)\\
   + \underset{x_{j} \in \mathcal{B} \, | \, y_{j} \in h(y_{i}) \cap \mathcal{L}}{\sum} g(f_{\theta}(x_{i}), f_{\theta}(x_{j}))\\
   + \underset{x_{j} \in \mathcal{B} \, | \, y_{j} \in \widetilde{\mathcal{L}}}{\sum} g(f_{\theta}(x_{i}), f_{\theta}(x_{j}))\end{matrix}}.
\end{equation}
Figure \ref{fig:overall_system}(b) shows a conceptualization of the proposed loss function.

\subsubsection{Hard Negative Sampling}
We utilize the modified loss function (\ref{eq:proxyncahn++}) by first training a base embedding model using (\ref{eq:proxynca++}) on OpenLogoDet3K47 in the closed-set regime and then computing its confusion matrix $C$ on the validation data, normalized along each row.
Then we construct the mapping $h:\mathcal{L} \to \mathbb{P}(\mathcal{L})$ by iterating over the rows $i$ of $C$, and setting $h(y_{i})$ to be the set of all classes $y_{j}$ such that $\alpha_{1} \leq C_{i,j} \leq \alpha_{2}$.
Through validation we determined $\alpha_{1} = 0.05$ and $\alpha_{2} = 0.35$ to avoid both choosing too-easy negatives and choosing classes $y_{j}$ that may be visually identical due to label noise.
As a secondary filter to avoid nearly identical classes, e.g. \textsf{atari\_1} versus \textsf{atari\_2}, we only kept $y_{j}$ if the levenshtein distance $l(y_{i}, y_{j}) > 2$.
Once we have constructed such a mapping $h$, we use a batch sampling strategy as follows.
For each batch we sample $k$ positive classes and $m$ samples per class.
Then for each class $l$ present, we choose the first random class $\widetilde{l} \in h(l)$ that is not already in the batch, if available.
We then sample an additional $m$ images with class $\widetilde{l}$, yielding a total batch size between $km$ and $2km$.
We investigate how this approach performs in Section \ref{subsec:ablation}.

Note that in the above scenario all hard negatives are themselves positives for some other class.
We could have also fed a large set of images through a logo-agnostic detector, generated embeddings for each detected region, and then mined the resulting regions sufficiently close to known positives for challenging negative examples that may not belong to any positive class, i.e. that belong to $\widetilde{\mathcal{L}}$.
However, this requires more annotation and is difficult to scale unless restricted to find hard negatives for a (small) subset of (challenging) classes.

\section{Experiments\label{sec:experiments}}
\subsection{Datasets}
There are numerous public datasets pertaining to logo recognition, and we mention those relevant to this work.
We also describe the construction of a combined dataset called OpenLogoDet3K47 that we used primarily for ablation studies (see Section \ref{subsec:ablation}). 
In general, we remove any images from these datasets with a minimum side length of less than $10$ pixels.
See Table \ref{table:datasets} for information on the main datasets considered.
When computing performance, we primarily use the recall@1 metric which is defined as the number of images in a query set identified correctly when retrieving their closest neighbors in a reference set.
We omit computing recall@k with $k > 1$ as in \cite{Li2022WACV} since visually similar logos often belong to very different types of brands due to trademark legislation, and thus practically, not retrieving the correct logo class via the closest neighbor would be considered a failure.
If a query image also exists in the reference set, then we take the second closest neighbor.

\subsubsection{QMUL-OpenLogo}
The QMUL-OpenLogo dataset \cite{OpenLogo2018} is itself constructed from 7 existing public datasets: FlickrLogos-27 \cite{FlickrLogos27},  FlickrLogos-32 \cite{Flickr322011}, Logos-32plus \cite{Logos32Plus2017}, BelgaLogos \cite{BelgaLogos09}, WebLogo-2M \cite{WebLogo2M2017}, Logos In The Wild \cite{LiTW2018}, and SportsLogo \cite{SportsLogo2017}.
The authors combined these datasets by combining some same-brand classes that exist in multiple sources, removing many erroneous annotations, and removing classes with fewer than $10$ images.
The result is a relatively high quality dataset containing $27189$ images across $309$ classes with object-level annotations.

When evaluating on this dataset with a model trained only on LogoDet3K, we closely follow \cite{Li2022WACV} and partition it into a query set, consisting of 10 random images per class, with the remaining images placed in a gallery set.
We also partition with respect to whether a class name contains ``-text" to provide an insight into performance on text-dominant logo classes.
Finally, we compute recall@1 over the entire dataset versus itself for comparison.

\subsubsection{LogoDet3K}
The LogoDet3K dataset \cite{LogoDet3K2022} consists of $158652$ images with $194261$ bounding boxes across $3000$ logo classes.
When training on this dataset, we use an open-set approach so that test classes are unseen.
Furthermore, we place any class with less than $20$ samples entirely in the training split.
When evaluating on this dataset, just as with OpenLogo we construct a query and gallery set.
However, this dataset does not specify text-dominant classes, so we cannot partition further.

\subsubsection{FlickrLogos-47}
FlickrLogos-47 \cite{Flickr322011} is an expanded version of FlickrLogos-32, where the same images are used but with more complete annotations that cover additional classes.

\subsubsection{OpenLogoDet3K47}
In order to improve the diversity of classes and volume of training data used in our experiments, we combine the LogoDet3K dataset with two other public datasets that have a relatively small overlap with it: QMUL-OpenLogo and FlickrLogos-47.
We cleaned the resulting union of datasets by merging or removing some redundant classes, and details can be found in the supplementary material. 
After removing image regions with a minimum side length less than $10$ pixels and any classes with fewer than $20$ instances, we have a total of $2714$ classes with $181552$ images and $227176$ objects.

\begin{figure}
\centering
\begin{tabular}{@{\hskip1pt}c@{\hskip1pt}c@{\hskip1pt}c@{\hskip1pt}c@{\hskip1pt}}
  \includegraphics[width=2cm,height=2cm]{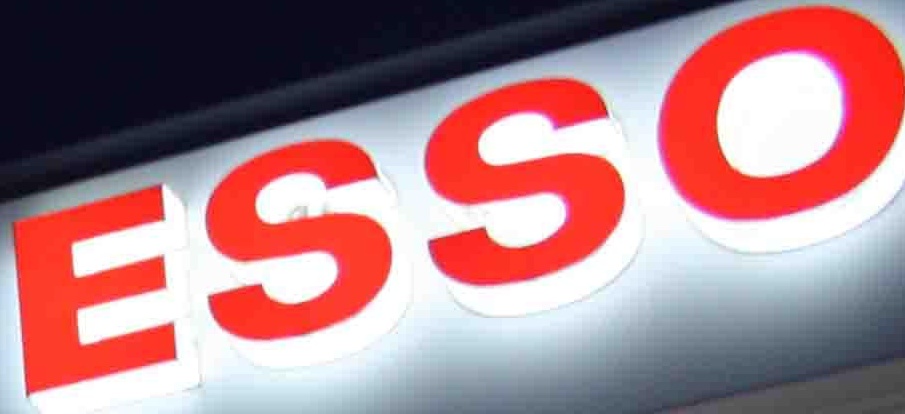} & \includegraphics[width=2cm,height=2cm]{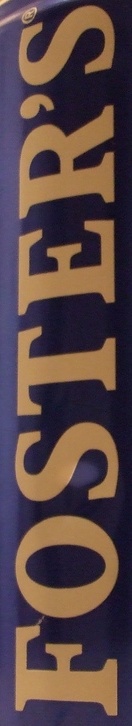} & \includegraphics[width=2cm,height=2cm]{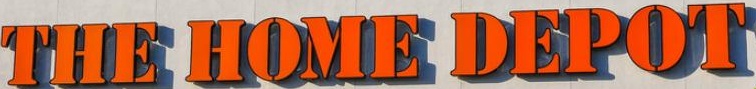} &   \includegraphics[width=2cm,height=2cm]{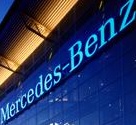}\\
  \textsf{\scriptsize esso\_text} & \textsf{\scriptsize fosters\_text} & \textsf{\scriptsize home\_depot\_text} & \textsf{\scriptsize mercedesbenz\_text} \\
  \includegraphics[width=2cm,height=2cm]{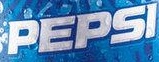} & \includegraphics[width=2cm,height=2cm]{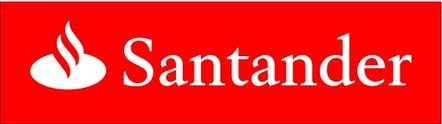} & \includegraphics[width=2cm,height=2cm]{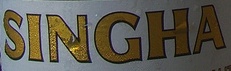} & \includegraphics[width=2cm,height=2cm]{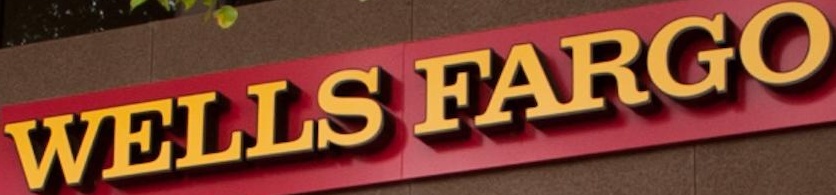} \\
  \textsf{\scriptsize pepsi\_text1} & \textsf{\scriptsize santander\_text} & \textsf{\scriptsize singha\_text} & \textsf{\scriptsize wellsfargo\_text} \\
\end{tabular}
\caption{Explicitly marked text-dominant classes from the test split of OpenLogoDet3K47}
\label{fig:testtextimages}
\end{figure}

Similar to OpenLogo and LogoDet3K, we evaluate on the test split by separating it into query and gallery subsets. 
In particular, we first split the dataset into a global query and gallery set.
We further split the query set into large images (greater than $70$ pixels on minimum side) and small images.
The value $70$ was chosen because it is near the $40^{th}$ percentile of minimum side length values across the entire dataset.
Finally, we separate small/large query subsets into those classes with ``\_text" in the name and those without. 
Figure \ref{fig:testtextimages} shows examples of such text-dominant classes from the test set.
This separation facilitates evaluation of recall@1 performance across specific parts of the test dataset, which in our evaluations include \textit{query vs gallery}, \textit{all vs all} (query plus gallery versus itself), \textit{small/large query vs gallery}, and \textit{text query vs gallery}.
Note that many classes without ``\_text" in the name are still text-heavy, so results must be interpreted carefully.

\subsubsection{E-commerce Image-Text Pair Dataset}
We curated approximately $20$ million image-text pairs from an e-commerce website, each containing a main product image and its corresponding product title.
This yields a similar dataset to WIT or Laion400m \cite{Laion2021} but in a different image domain.
We then trained a CLIP model on this dataset to obtain an alternative image-text pre-trained ViT base model to fine-tune for logo recognition.
\begin{table*}
  \begin{center}
  \caption{Public Logo Dataset Comparison}
  \label{table:datasets}
  \scriptsize
  \begin{tabular}{lcccll}
  \hline
  \noalign{\smallskip}
  Dataset & Logos & Images & Objects & Annotation & Text/Sym. Sep. \\
  \noalign{\smallskip}
  \hline
  BelgaLogos \cite{BelgaLogos09} & 37 & 1321 & 2697 & Object-Level & \checkmark \\
  \hline
  LogosInTheWild \cite{LiTW2018} & 871 (786) & 11054 & 32850 (25311) & Mixed & \checkmark \\
  \hline
  FlickrLogo-32 \cite{Flickr322011} & 32 & 2240 & 3404 & Object-Level & \checkmark \\
  \hline
  FlickrLogo-47 \cite{Flickr322011} & 47 & 2235 & 5968 & Object-Level & \checkmark \\
  \hline
  QMUL-OpenLogo \cite{OpenLogo2018} & 352 & 27083 & 51207 & Object-Level & \checkmark \\
  \hline
  LogoDet3K \cite{LogoDet3K2022} & 3000 & 158652 & 194261 & Object-Level & \\
  \hline
  OpenLogoDet3K47 & 3276 (2714) & 188244 (181552) & 235738 (227176) & Object-Level & Partial \\
  \hline
  \end{tabular}
  \end{center}
\end{table*}

\subsection{Implementation}
All experiments were conducted on machines with $4$ Tesla V100 GPUs, using PyTorch version 1.10.
The training code was built on the Pytorch Metric Learning Library, version 0.9.99.
Three different architectures were used: a ViT with patch size $32$ (ViT/B-32 \cite{Clip2021}); a CLIP-modified ResNet50 architecture (CLIP-ResNet50) whose final pooling is replaced with multi-head attention; a standard ResNet50 architecture from \textsf{torchvision} with a modified final pooling and layer norm before output \cite{Torchvision2010,Teh2020}.

The output embedding has $128$ dimensions for all models, which we obtain by attaching a fully connected (FC) layer of shape $128 \times d_{0}$, where $d_{0}$ is the output dimension of the base (trunk) architecture.
We initialize the proxy vectors by evaluating the pre-trained model (with added fully connected layer) on the training set and computing per-class mean embeddings.
All networks were adapted to process input images with $160 \times 160$ pixels, which is comparable with input image sizes used in \cite{Li2022WACV,Istvan2019}.
This is accomplished for ResNet50 by replacing the final average pool layer with an adaptive max pooling layer.
For the ViT and modified ResNet50 models used by CLIP, we apply bicubic interpolation to the positional embedding tensors.

Common hyperparameters used are influenced by referencing those cited in \cite{Clip2021,Teh2020}.
Hyperparameter tuning was conducted on a validation set with approximately $16\%$ of logo classes.
The training set comprises approximately $64\%$ of classes, with the remaining $20\%$ of classes assigned to the test set.
Any collected negative samples with class in $h(y) \cap \mathcal{L}$ were included in the split containing all positives for class $y$.
For each model architecture and pre-trained weights considered, we conducted $20$ trials using the Ax library to select the trunk base learning rate $\gamma_{\mathrm{trunk}} \in [10^{-6}, 10^{-4}]$, final fully connected layer learning rate $\gamma_{\mathrm{fc}} \in [0.0001, 0.01]$, proxy learning rate $\gamma_{p} \in \{1,\, 2,\, \ldots, \, 100\}$, and batch size $b \in \{128, 192\}$.
Each trial was trained for $25$ epochs using the AdamW optimizer with learning rates reduced on plateau (patience of $4$ and factor of $0.25$).
The temperature scaling parameter $\sigma$ in (\ref{eq:proxynca++}) and (\ref{eq:proxyncahn++}) was fixed at $0.06$.
Image augmentation used consisted of color jitter and perspective transformations.
Other fixed hyperparameters are listed in the supplementary material.
The best performing hyperparameter set for each model configuration was used to train a final model on the training and validation sets combined, with evaluation metrics then computed on the test set.
In order to build a ViT pre-trained on E-commerce image-text data, we used the same hyperparameters as specified in \cite{Clip2021}.
We randomly initialize the vision-text backbone of the CLIP model based on ViT and train with a contrastive loss.

\subsection{Comparison With State-Of-Art On Public Datasets\label{subsec:publiccomparison}}
In Table \ref{table:logodet3kresults}, we see the performance of our best image-text pre-trained model on several unseen public datasets, compared to the previous state-of-art \cite{Li2022WACV} that was also trained on the LogoDet3K dataset.
Our model uses $160$ pixel input size compared to $150$ for the comparison model, and we use $128$-dimensional embeddings instead of $512$ ($1024$ with combined text recognition and image features).
We trained on the same percentage of logo classes from LogoDet3K as in \cite{Li2022WACV} (80\%), but there are no standard splits for this dataset, so we could only ensure that we sampled and prepared the data as closely as possible.
In the case of OpenLogo and the LogoDet3K test set, we again mimic \cite{Li2022WACV} and split each into query and gallery sets to give more realistic estimates of zero-shot performance.
We show text-specific recall@1 where possible.
Ultimately, we see a roughly $3$-$6\%$ improvement in recall@1 across all subsets where exact comparison data is available, except for BelgaLogo, which can likely be attributed to its small size.
\begin{table*}
   \begin{center}
   \caption{Recall@1 performance on various public logo datasets for the best ViT base model trained on LogoDet3K vs previous state-of-art taken from Table 4 of \cite{Li2022WACV}.}
   \label{table:logodet3kresults}
   \small
   \begin{tabular}{lccccccccccc}
   \hline
   Model & \multicolumn{2}{c}{LogoDet3K Test} &  \multicolumn{3}{c}{OpenLogo} & \multicolumn{2}{c}{BelgaLogo} & \multicolumn{2}{c}{FlickrLogos-47} & \multicolumn{2}{c}{LiTW} \\
         &  Q vs G & All & Q vs G & All & Text & All & Text & All & Text & All & Text \\
   \hline
   ViT IT Pre-trained & \bf{0.9836} & 0.9886 & \bf{0.9371} & 0.9629 & 0.9568 & \bf{0.9797} & 0.9784 & \bf{0.9834} & \bf{0.9778} & \bf{0.9391} & 0.9456\\
   SeeTek \cite{Li2022WACV} & 0.9490 & - & 0.8964 & - & - & 0.9739 & \bf{1.0} & 0.9188 & 0.9400 & 0.8772 & -\\
\hline
\end{tabular}
\end{center}
\end{table*}

\subsection{Ablation Study\label{subsec:ablation}}
\begin{figure}
\centering
\begin{tabular}{@{\hskip1pt}c@{\hskip1pt}@{\hskip1pt}c@{\hskip1pt}@{\hskip1pt}c@{\hskip1pt}@{\hskip1pt}c@{\hskip1pt}@{\hskip1pt}c@{\hskip1pt}@{\hskip1pt}c@{\hskip1pt}}
  \textsf{\small Query} & \textsf{\small Incorrect} & \textsf{\small Correct} & \textsf{\small Query} & \textsf{\small Incorrect} & \textsf{\small Correct}\\
  \includegraphics[width=1cm,height=1cm]{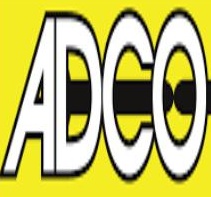} & \includegraphics[width=1cm,height=1cm]{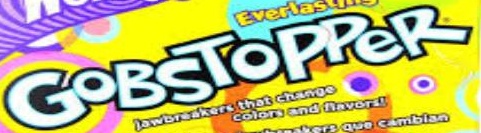} & \includegraphics[width=1cm,height=1cm]{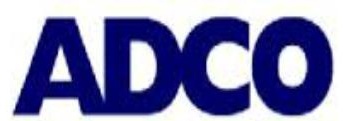} & \includegraphics[width=1cm,height=1cm]{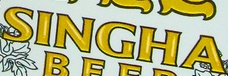} & \includegraphics[width=1cm,height=1cm]{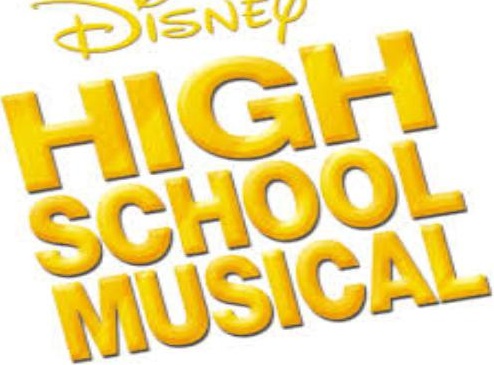} & \includegraphics[width=1cm,height=1cm]{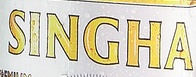}\\
  \includegraphics[width=1cm,height=1cm]{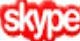} & \includegraphics[width=1cm,height=1cm]{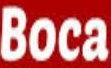} & \includegraphics[width=1cm,height=1cm]{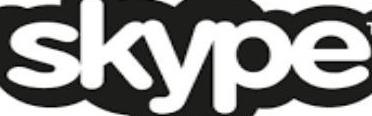} & \includegraphics[width=1cm,height=1cm]{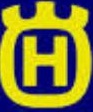} & \includegraphics[width=1cm,height=1cm]{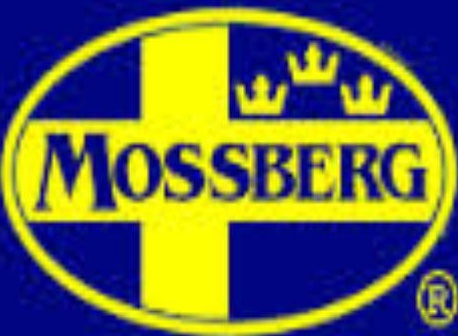} & \includegraphics[width=1cm,height=1cm]{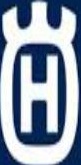}\\
  \includegraphics[width=1cm,height=1cm]{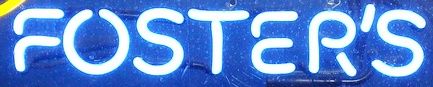} & \includegraphics[width=1cm,height=1cm]{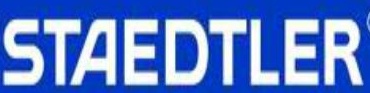} & \includegraphics[width=1cm,height=1cm]{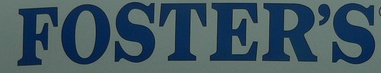} &     \includegraphics[width=1cm,height=1cm]{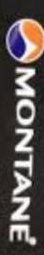} & \includegraphics[width=1cm,height=1cm]{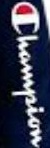} & \includegraphics[width=1cm,height=1cm]{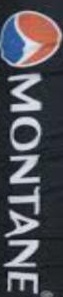} \\
  \includegraphics[width=1cm,height=1cm]{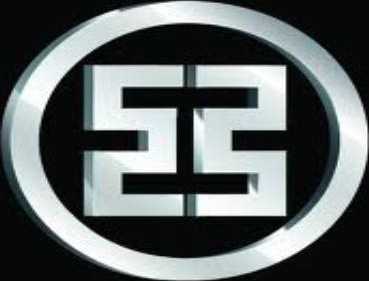} & \includegraphics[width=1cm,height=1cm]{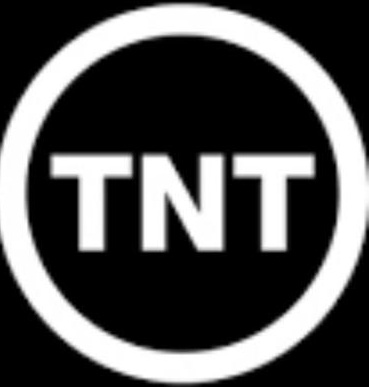} & \includegraphics[width=1cm,height=1cm]{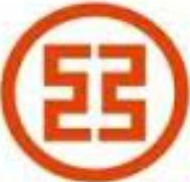} & \includegraphics[width=1cm,height=1cm]{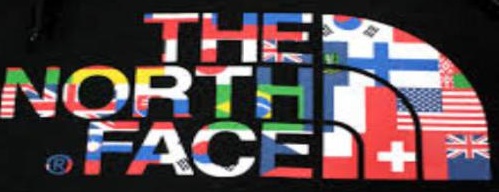} & \includegraphics[width=1cm,height=1cm]{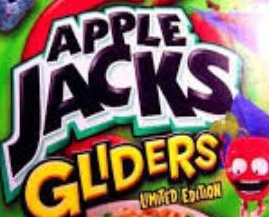} & \includegraphics[width=1cm,height=1cm]{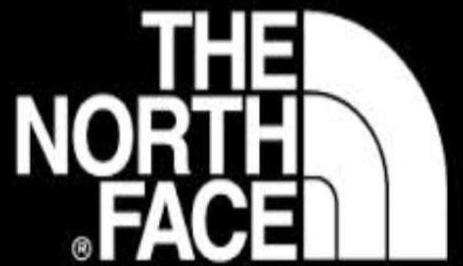}\\
  \includegraphics[width=1cm,height=1cm]{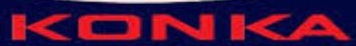} & \includegraphics[width=1cm,height=1cm]{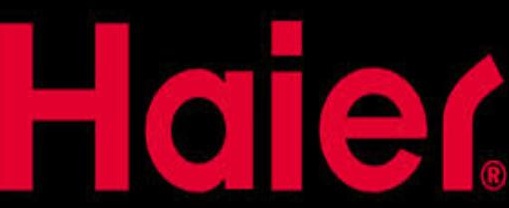} & \includegraphics[width=1cm,height=1cm]{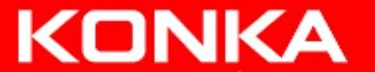} & \includegraphics[width=1cm,height=1cm]{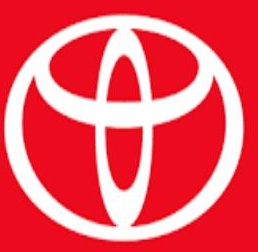} & \includegraphics[width=1cm,height=1cm]{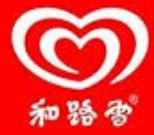} & \includegraphics[width=1cm,height=1cm]{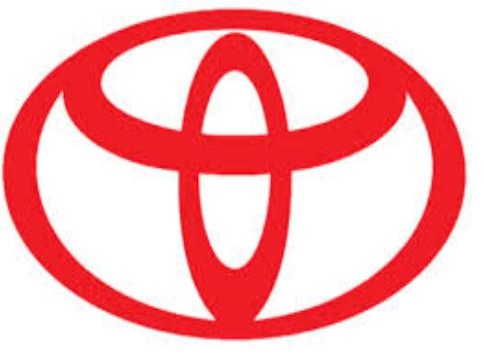}\\
  \includegraphics[width=1cm,height=1cm]{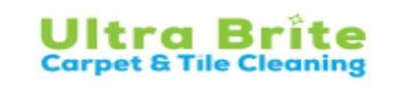} & \includegraphics[width=1cm,height=1cm]{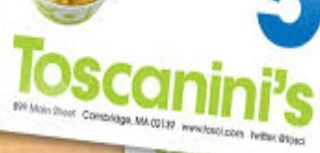} & \includegraphics[width=1cm,height=1cm]{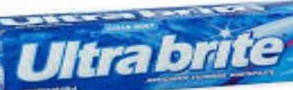} &     \includegraphics[width=1cm,height=1cm]{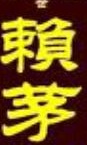} & \includegraphics[width=1cm,height=1cm]{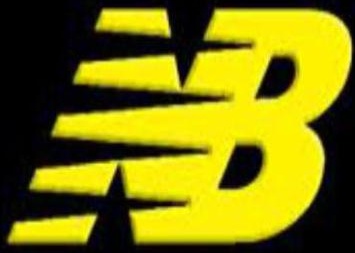} & \includegraphics[width=1cm,height=1cm]{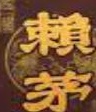}\\
  \includegraphics[width=1cm,height=1cm]{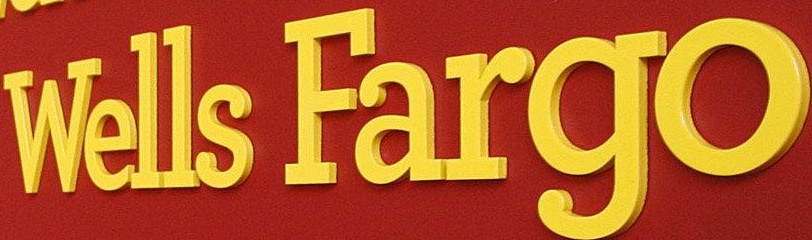} & \includegraphics[width=1cm,height=1cm]{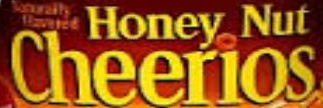} & \includegraphics[width=1cm,height=1cm]{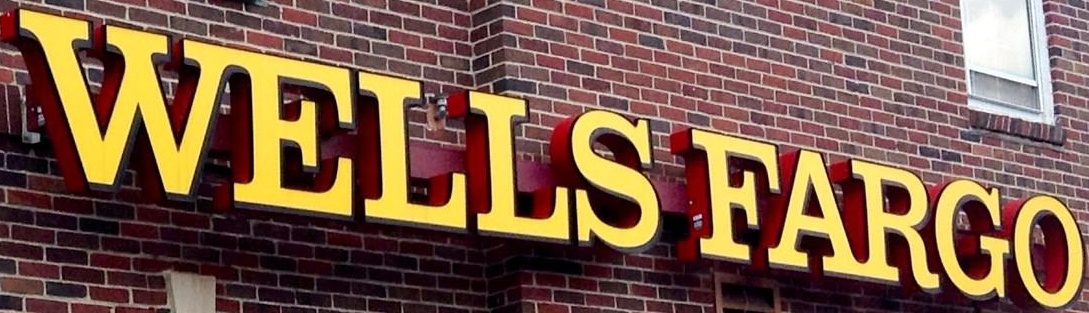} & \includegraphics[width=1cm,height=1cm]{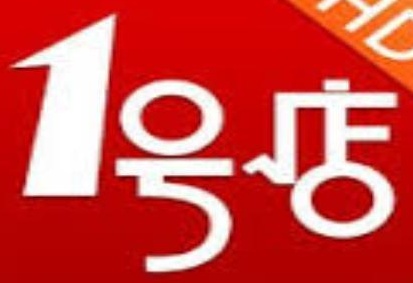} & \includegraphics[width=1cm,height=1cm]{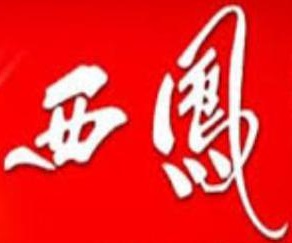} & \includegraphics[width=1cm,height=1cm]{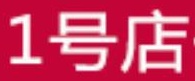}\\
  \includegraphics[width=1cm,height=1cm]{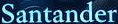} & \includegraphics[width=1cm,height=1cm]{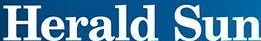} & \includegraphics[width=1cm,height=1cm]{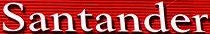} & \includegraphics[width=1cm,height=1cm]{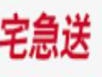} & \includegraphics[width=1cm,height=1cm]{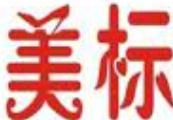} & \includegraphics[width=1cm,height=1cm]{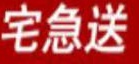}
\end{tabular}
\caption{Sample images from the OpenLogoDet3K47 dataset test split showing that our image-text pre-trained ViT logo embedder is better at representing text-dominant logo classes when compared to the same architecture pre-trained on ImageNet. Columns 1+4: query image, columns 2+5: ImageNet-pre-trained ViT model's incorrect retrieval, columns 3+6: Image-text pre-trained ViT model's correct logo retrieval}
\label{fig:qualcomp}
\end{figure}

We now consider how architecture and pre-training method affect overall performance.
We observe in Table \ref{table:3k47ablation} that both image-text pre-trained models (either from OpenAI or trained on E-commerce image-title pairs) yield nearly 1\% better recall@1 performance (test set queried against itself) than either ImageNet1K-pre-trained model.
Furthermore, the recall@1 performance with queries restricted to text-dominant classes has a 2-3\% improvement.
We show in Figure \ref{fig:qualcomp} several failure cases of the best ImageNet1K-pre-trained ViT model relative to the best OpenAI image-text pre-trained ViT model.
In particular, we see that the image-text pre-trained model retrieves the correct logo for text-heavy classes like `Fosters', `Konka', or `Ultra Brite', where the font style/color/background differs, while the ImageNet pre-trained retrieves an incorrect match.
We also see evidence of OCR sensitivity to non-English (Chinese) characters as evidenced by correct matches in rows 6-8 of Figure \ref{fig:qualcomp}.
\begin{table*}[!ht]
\begin{center}
\caption{Recall@1 performance on the OpenLogoDet3K47 test split among several model initializations.
\textit{Pre-training} denotes which data the base model was pre-trained on.
For example, \textit{OpenAI IT} refers to OpenAI's image-text pre-trained ViT-B/32 model.
\textit{BS} is batch size; \textit{LR (T, E, P)} denotes the trunk, embedder, and proxy learning rates, respectively; \textit{All vs All} denotes recall@1 when querying the entire test dataset against itself and taking the 2nd closest neighbor; \textit{Q vs G} refers to recall@1 for disjoint query and gallery subsets of the test set.}
\label{table:3k47ablation}
\small
\begin{tabular}{llcccccccc}
\hline
\noalign{\smallskip}
Model & Pre-training & BS & LR & NMI & All vs All & Q vs G & Text & Small & Large \\
      &             &    & (T,E,P) & &        &     &      & (QvG) & (QvG) \\
\noalign{\smallskip}
\hline
ViT-B/32  & OpenAI IT & 192 & (2.3E-06, 0.0015, 71) & \bf{0.9095} & \bf{0.9867} & \bf{0.9788} & \bf{0.9925} & \bf{0.9702} & \bf{0.9843}\\ 
\hline
ViT-B/32 & E-comm. IT & 192 & (5.0E-06, 0.0015, 70) & 0.9040 & 0.9849 & 0.9759 & 0.9888 & 0.9664 & 0.9819 \\
\hline
CLIP-ResNet50 & OpenAI IT & 128 & (5.3E-06, 0.0005, 67) & 0.8854 & 0.9783 & 0.9641 & 0.9716 & 0.9518 & 0.9720 \\
\hline
ResNet50& ImageNet1K & 128 & (0.00011, 0.00045, 54) & 0.8742 & 0.9756 & 0.9551 & 0.9765 & 0.9518 & 0.9573\\
\hline
ViT-B/32 & ImageNet1K & 128 & (3.3E-05, 0.0023, 100) & 0.8707 & 0.9755 & 0.9523 & 0.9790 & 0.9504 & 0.9536 \\
\hline
\end{tabular}
\end{center}
\end{table*}

Regarding the effect of loss function (\ref{eq:proxyncahn++}), we observe in Table \ref{table:3k47hnablation} that overall recall@1 improves slightly by $0.1\%$.
To see this we trained two models in the closed-set setting (i.e. all classes seen during training) using (\ref{eq:proxynca++}) or (\ref{eq:proxyncahn++}), while holding all other hyperparameters the same.
\begin{table*}[!ht]
  \begin{center}
  \caption{Recall@1 performance on the test split of OpenLogoDet3K47 dataset using loss function (\ref{eq:proxynca++}) vs (\ref{eq:proxyncahn++}).
  The model was generated by fine-tuning the OpenAI image-text pre-trained ViT model on the (closed-set) training and validation splits of OpenLogoDet3K47 (i.e. all classes are seen during training)}
  \label{table:3k47hnablation}
  \small
  \begin{tabular}{l|l|c|c|c|c|c|c|c|c|c}
  \hline
  \noalign{\smallskip}
    Model & Pretraining & BS & LR     & Loss & NMI & All vs All & Q vs G & Text  & Small & Large \\
          &             &    & (T,E,P) &    &      &         &     &        & (QvG) & (QvG) \\
  \noalign{\smallskip}
  \hline
  ViT-B/32 & OpenAI IT & 192 & (3E-06, 0.0003, 63) & (\ref{eq:proxynca++}) & 0.9325 & 0.9588 & \bf{0.9480} & 0.9341 & \bf{0.9340} & 0.9583\\
  ViT-B/32 & OpenAI IT & 192 & (3E-06, 0.0003, 63) & (\ref{eq:proxyncahn++}) & \bf{0.9339} & \bf{0.9600} & 0.9475 & \bf{0.9366} & 0.9312 & \bf{0.9594}\\
  \hline
  \end{tabular}
  \end{center}
\end{table*}

\begin{figure}[!ht]
  \centering
  \begin{tabular}{@{\hskip1pt}c@{\hskip1pt}c@{\hskip1pt}}
    \includegraphics[width=4.25cm]{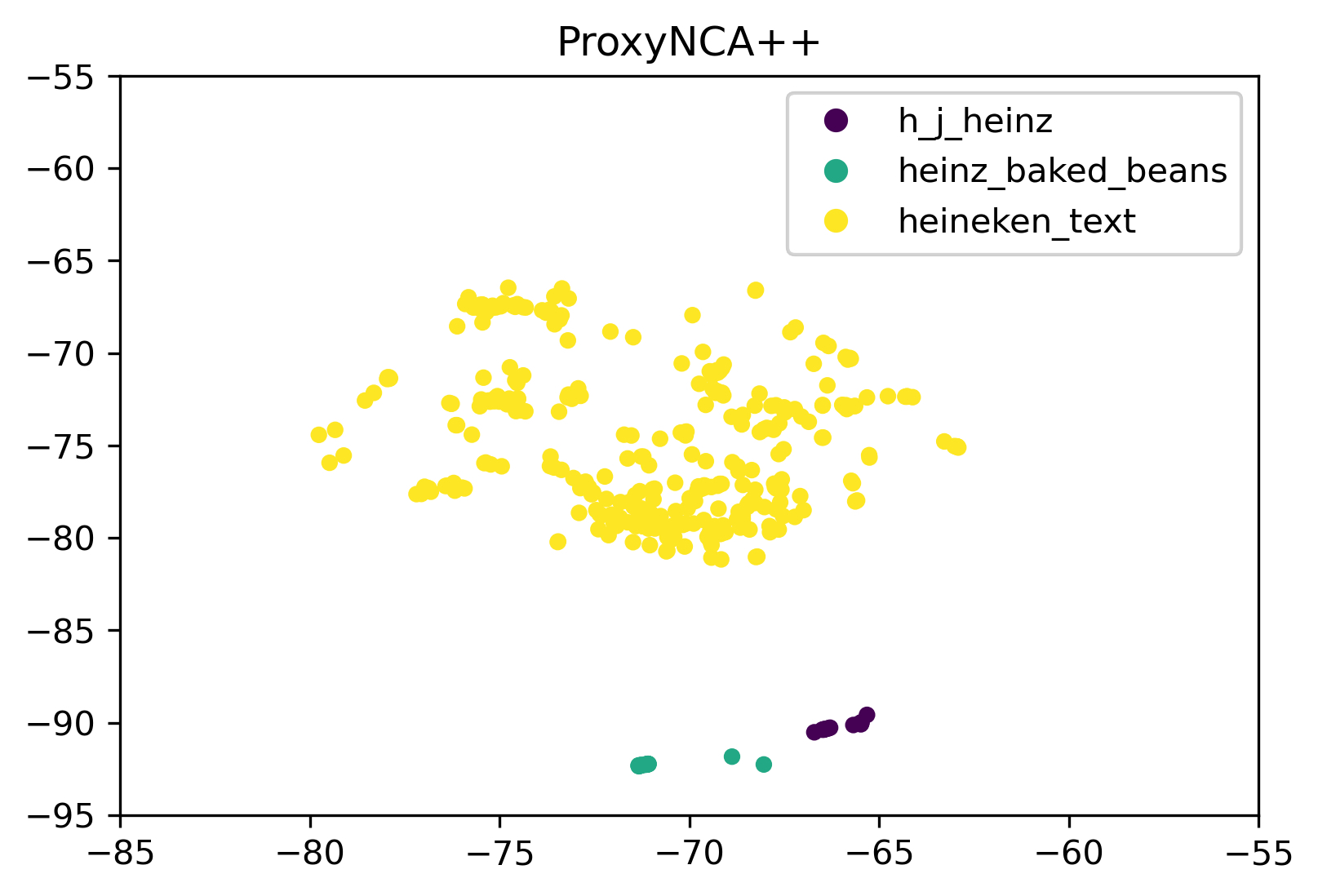} & \includegraphics[width=4.25cm]{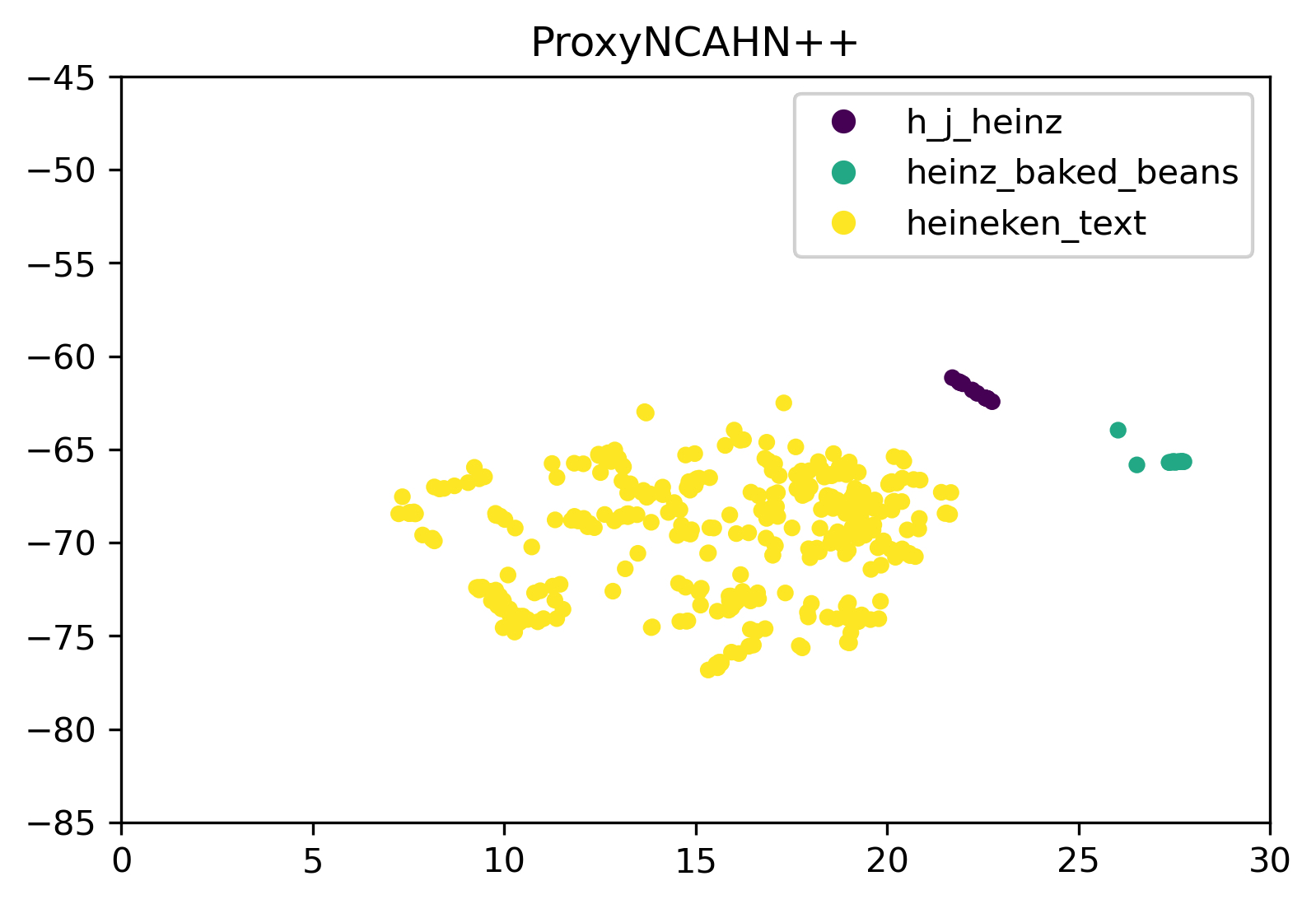}
  \end{tabular}
  \caption{TSNE comparison for test images from hard negative cohort $\{\textsf{h\_j\_heinz}, \textsf{heinz\_baked\_beans}, \textsf{heineken\_text}\}$ using models trained with ProxyNCA++ versus ProxyNCAHN++ loss functions.}
  \label{fig:tsne}
\end{figure}
Furthermore, looking at the hard negative set mapping $\textsf{h\_j\_heinz} \to h(\textsf{h\_j\_heinz}) = \{\textsf{heinz\_baked\_beans}, \textsf{heineken\_text}\}$, we observe that the corresponding normalized confusion submatrix slightly improves as shown in (\ref{eq:cm_change}).
Also, in Figure \ref{fig:tsne} the model trained with (\ref{eq:proxyncahn++}) has qualitatively better separation of \textsf{h\_j\_heinz} from \textsf{heinz\_baked\_beans}.
\begin{equation}
\label{eq:cm_change}
\setlength\arraycolsep{1pt}
\begin{array}{r}
  \textsf{\tiny{h\_j\_heinz}}\\
  \textsf{\tiny{heinz\_baked\_beans}}\\
  \textsf{\tiny{heineken\_text}}
\end{array}
\begin{pmatrix}
1 & 0 & 0\\
0 & 0.909 & 0.045\\
0 & 0 & 0.979
\end{pmatrix} \longrightarrow \begin{pmatrix}
   1 & 0 & 0\\
   0 & 0.954 & 0\\
   0 & 0 & 0.966
\end{pmatrix}.
\end{equation}

\section{Conclusion}
\begin{figure}
  \centering
  \begin{tabular}{@{\hskip1pt}c@{\hskip1pt}@{\hskip1pt}c@{\hskip1pt}@{\hskip1pt}c@{\hskip1pt}@{\hskip1pt}c@{\hskip1pt}@{\hskip1pt}c@{\hskip1pt}@{\hskip1pt}c@{\hskip1pt}}
    \textsf{\small Query} & \textsf{\small Retrieval} & \textsf{\small Query} & \textsf{\small Retrieval} & \textsf{\small Query} & \textsf{\small Retrieval}\\
    \includegraphics[width=1cm,height=1cm]{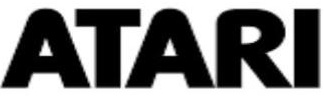} & \includegraphics[width=1.25cm,height=1.25cm]{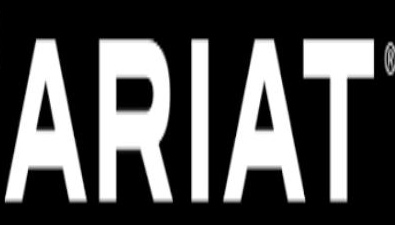} & \includegraphics[width=1.25cm,height=1.25cm]{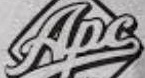} & \includegraphics[width=1.25cm,height=1.25cm]{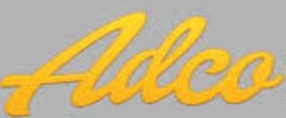} & \includegraphics[width=1.25cm,height=1.25cm]{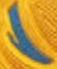} & \includegraphics[width=1.25cm,height=1.25cm]{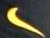} \\
    \includegraphics[width=1.25cm,height=1.25cm]{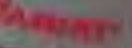} & \includegraphics[width=1.25cm,height=1.25cm]{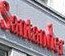} & \includegraphics[width=1.25cm,height=1.25cm]{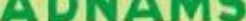} &     \includegraphics[width=1.25cm,height=1.25cm]{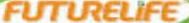} & \includegraphics[width=1.25cm,height=1.25cm]{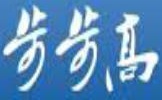} & \includegraphics[width=1.25cm,height=1.25cm]{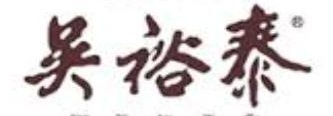} \\
    \includegraphics[width=1.25cm,height=1.25cm]{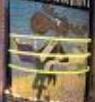} & \includegraphics[width=1.25cm,height=1.25cm]{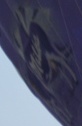} & \includegraphics[width=1.25cm,height=1.25cm]{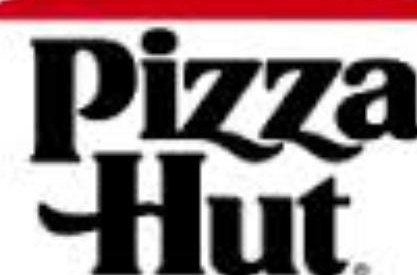} & \includegraphics[width=1.25cm,height=1.25cm]{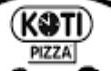} & \includegraphics[width=1.25cm,height=1.25cm]{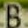} & \includegraphics[width=1.25cm,height=1.25cm]{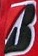} 
  \end{tabular}
  \caption{Logo retrieval errors showing different challenges for future work}
  \label{fig:futurechallenges}
  \end{figure}

This work demonstrated a simple yet effective solution to embue text sensitivity in a logo matching model, which sets new state-of-the-art on all public logo datasets.
Our solution leverages image-text multimodal pre-training to discover OCR capabilities which proves vital on text-heavy logo images.
Notably, the same model continues to outperform the existing state-of-the-art matching model even on non-text logos.
The new proposed ProxyNCAHN++ loss function improves separability on top of a strong baseline.
In the future, we will study a better integration of the logo-agnostic detector to understand how upstream detection accuracy influences downstream precision.
Further challenges also remain regarding poorly aligned logo bounding boxes in the context of a logo detector/embedder pipeline, blurry/tiny logo regions, poor lighting conditions, partial text matches, ambiquous or indistinct logos, and stylized or non-English text (see Figure \ref{fig:futurechallenges}).

{\small
\bibliographystyle{ieee_fullname}
\bibliography{references}

\begin{thebibliography}{10}\itemsep=-1pt

\bibitem{Bastan2022}
Muhammet Bastan, Hao-Yu Wu, Tian Cao, Bhargava Kota, and Mehmet Tek.
\newblock Large scale open-set deep logo detection, 2022.

\bibitem{Logos32Plus2017}
Simone Bianco, Marco Buzzelli, Davide Mazzini, and Raimondo Schettini.
\newblock Deep learning for logo recognition.
\newblock {\em Neurocomput.}, 245(C):23–30, jul 2017.

\bibitem{YoloV42020}
Alexey Bochkovskiy, Chien-Yao Wang, and Hong-Yuan~Mark Liao.
\newblock Yolov4: Optimal speed and accuracy of object detection, 2020.

\bibitem{Cesarini}
F. Cesarini, E. Francesconi, M. Gori, S. Marinai, J.Q. Sheng, and G. Soda.
\newblock A neural-based architecture for spot-noisy logo recognition.
\newblock In {\em Proceedings of the Fourth International Conference on
  Document Analysis and Recognition}, volume~1, pages 175--179 vol.1, 1997.

\bibitem{deng2009imagenet}
Jia Deng, Wei Dong, Richard Socher, Li-Jia Li, Kai Li, and Li Fei-Fei.
\newblock Imagenet: A large-scale hierarchical image database.
\newblock In {\em 2009 IEEE conference on computer vision and pattern
  recognition}, pages 248--255. Ieee, 2009.

\bibitem{devlin2018bert}
Jacob Devlin, Ming-Wei Chang, Kenton Lee, and Kristina Toutanova.
\newblock Bert: Pre-training of deep bidirectional transformers for language
  understanding.
\newblock {\em arXiv preprint arXiv:1810.04805}, 2018.

\bibitem{doermann}
D.S. Doermann, E. Rivlin, and I. Weiss.
\newblock Logo recognition using geometric invariants.
\newblock In {\em Proceedings of 2nd International Conference on Document
  Analysis and Recognition (ICDAR '93)}, pages 894--897, 1993.

\bibitem{dosovitskiy2020image}
Alexey Dosovitskiy, Lucas Beyer, Alexander Kolesnikov, Dirk Weissenborn,
  Xiaohua Zhai, Thomas Unterthiner, Mostafa Dehghani, Matthias Minderer, Georg
  Heigold, Sylvain Gelly, et~al.
\newblock An image is worth 16x16 words: Transformers for image recognition at
  scale.
\newblock {\em arXiv preprint arXiv:2010.11929}, 2020.

\bibitem{Istvan2019}
István Fehérvári and Srikar Appalaraju.
\newblock Scalable logo recognition using proxies.
\newblock In {\em 2019 IEEE Winter Conference on Applications of Computer
  Vision (WACV)}, pages 715--725, 2019.

\bibitem{girshick2015fast}
Ross Girshick.
\newblock Fast r-cnn.
\newblock In {\em Proceedings of the IEEE international conference on computer
  vision}, pages 1440--1448, 2015.

\bibitem{hadsell2006dimensionality}
Raia Hadsell, Sumit Chopra, and Yann LeCun.
\newblock Dimensionality reduction by learning an invariant mapping.
\newblock In {\em 2006 IEEE Computer Society Conference on Computer Vision and
  Pattern Recognition (CVPR'06)}, volume~2, pages 1735--1742. IEEE, 2006.

\bibitem{ResNet2016}
Kaiming He, Xiangyu Zhang, Shaoqing Ren, and Jian Sun.
\newblock Deep residual learning for image recognition.
\newblock In {\em 2016 IEEE Conference on Computer Vision and Pattern
  Recognition (CVPR)}, pages 770--778, 2016.

\bibitem{hu2020multimodal}
Changbo Hu, Qun Li, Zhen Zhang, Keng-hao Chang, and Ruofei Zhang.
\newblock A multimodal fusion framework for brand recognition from product
  image and context.
\newblock In {\em 2020 IEEE International Conference on Multimedia \& Expo
  Workshops (ICMEW)}, pages 1--4. IEEE, 2020.

\bibitem{jia2021scaling}
Chao Jia, Yinfei Yang, Ye Xia, Yi-Ting Chen, Zarana Parekh, Hieu Pham, Quoc Le,
  Yun-Hsuan Sung, Zhen Li, and Tom Duerig.
\newblock Scaling up visual and vision-language representation learning with
  noisy text supervision.
\newblock In {\em International Conference on Machine Learning}, pages
  4904--4916. PMLR, 2021.

\bibitem{BelgaLogos09}
Alexis Joly and Olivier Buisson.
\newblock Logo retrieval with a contrario visual query expansion.
\newblock In {\em MM '09: Proceedings of the seventeen ACM international
  conference on Multimedia}, pages 581--584, 2009.

\bibitem{FlickrLogos27}
Y. Kalantidis, LG. Pueyo, M. Trevisiol, R. van Zwol, and Y. Avrithis.
\newblock Scalable triangulation-based logo recognition.
\newblock In {\em in Proceedings of ACM International Conference on Multimedia
  Retrieval (ICMR 2011)}, Trento, Italy, April 2011.

\bibitem{kant2020spatially}
Yash Kant, Dhruv Batra, Peter Anderson, Alexander Schwing, Devi Parikh, Jiasen
  Lu, and Harsh Agrawal.
\newblock Spatially aware multimodal transformers for textvqa.
\newblock In {\em European Conference on Computer Vision}, pages 715--732.
  Springer, 2020.

\bibitem{kiela2020hateful}
Douwe Kiela, Hamed Firooz, Aravind Mohan, Vedanuj Goswami, Amanpreet Singh,
  Pratik Ringshia, and Davide Testuggine.
\newblock The hateful memes challenge: Detecting hate speech in multimodal
  memes.
\newblock {\em Advances in Neural Information Processing Systems},
  33:2611--2624, 2020.

\bibitem{kim2021vilt}
Wonjae Kim, Bokyung Son, and Ildoo Kim.
\newblock Vilt: Vision-and-language transformer without convolution or region
  supervision.
\newblock In {\em International Conference on Machine Learning}, pages
  5583--5594. PMLR, 2021.

\bibitem{kingma2014adam}
Diederik~P Kingma and Jimmy Ba.
\newblock Adam: A method for stochastic optimization.
\newblock {\em arXiv preprint arXiv:1412.6980}, 2014.

\bibitem{Li2022WACV}
Chenge Li, Istv\'an Feh\'erv\'ari, Xiaonan Zhao, Ives Macedo, and Srikar
  Appalaraju.
\newblock Seetek: Very large-scale open-set logo recognition with text-aware
  metric learning.
\newblock In {\em Proceedings of the IEEE/CVF Winter Conference on Applications
  of Computer Vision (WACV)}, pages 2544--2553, January 2022.

\bibitem{li2021align}
Junnan Li, Ramprasaath Selvaraju, Akhilesh Gotmare, Shafiq Joty, Caiming Xiong,
  and Steven Chu~Hong Hoi.
\newblock Align before fuse: Vision and language representation learning with
  momentum distillation.
\newblock {\em Advances in Neural Information Processing Systems}, 34, 2021.

\bibitem{SportsLogo2017}
Yuan Liao, Xiaoqing Lu, Chengcui Zhang, Yongtao Wang, and Zhi Tang.
\newblock Mutual enhancement for detection of multiple logos in sports videos.
\newblock In {\em 2017 IEEE International Conference on Computer Vision
  (ICCV)}, pages 4856--4865, 2017.

\bibitem{lowe1999object}
David~G Lowe.
\newblock Object recognition from local scale-invariant features.
\newblock In {\em Proceedings of the seventh IEEE international conference on
  computer vision}, volume~2, pages 1150--1157. Ieee, 1999.

\bibitem{Torchvision2010}
S{\'e}bastien Marcel and Yann Rodriguez.
\newblock Torchvision the machine-vision package of torch.
\newblock {\em Proceedings of the 18th ACM international conference on
  Multimedia}, 2010.

\bibitem{ProxyNCA2017}
Y. Movshovitz-Attias, A. Toshev, T.~K. Leung, S. Ioffe, and S. Singh.
\newblock No fuss distance metric learning using proxies.
\newblock In {\em 2017 IEEE International Conference on Computer Vision
  (ICCV)}, pages 360--368, Los Alamitos, CA, USA, oct 2017. IEEE Computer
  Society.

\bibitem{Clip2021}
Alec Radford, Jong~Wook Kim, Chris Hallacy, Aditya Ramesh, Gabriel Goh,
  Sandhini Agarwal, Girish Sastry, Amanda Askell, Pamela Mishkin, Jack Clark,
  Gretchen Krueger, and Ilya Sutskever.
\newblock Learning transferable visual models from natural language
  supervision.
\newblock In Marina Meila and Tong Zhang, editors, {\em Proceedings of the 38th
  International Conference on Machine Learning}, volume 139 of {\em Proceedings
  of Machine Learning Research}, pages 8748--8763. PMLR, 18--24 Jul 2021.

\bibitem{redmon2017yolo9000}
Joseph Redmon and Ali Farhadi.
\newblock Yolo9000: better, faster, stronger.
\newblock In {\em Proceedings of the IEEE conference on computer vision and
  pattern recognition}, pages 7263--7271, 2017.

\bibitem{ren2015faster}
Shaoqing Ren, Kaiming He, Ross Girshick, and Jian Sun.
\newblock Faster r-cnn: Towards real-time object detection with region proposal
  networks.
\newblock {\em Advances in neural information processing systems}, 28, 2015.

\bibitem{Flickr322011}
Stefan Romberg, Lluis~Garcia Pueyo, Rainer Lienhart, and Roelof van Zwol.
\newblock Scalable logo recognition in real-world images.
\newblock In {\em Proceedings of the 1st ACM International Conference on
  Multimedia Retrieval}, ICMR '11, New York, NY, USA, 2011. Association for
  Computing Machinery.

\bibitem{tripletloss}
Florian Schroff, Dmitry Kalenichenko, and James Philbin.
\newblock Facenet: A unified embedding for face recognition and clustering.
\newblock In {\em 2015 IEEE Conference on Computer Vision and Pattern
  Recognition (CVPR)}, pages 815--823, 2015.

\bibitem{Laion2021}
Christoph Schuhmann, Richard Vencu, Romain Beaumont, Robert Kaczmarczyk,
  Clayton Mullis, Aarush Katta, Theo Coombes, Jenia Jitsev, and Aran
  Komatsuzaki.
\newblock Laion-400m: Open dataset of clip-filtered 400 million image-text
  pairs, 2021.

\bibitem{WebLogo2M2017}
Hang Su, Shaogang Gong, and Xiatian Zhu.
\newblock Weblogo-2m: Scalable logo detection by deep learning from the web.
\newblock In {\em 2017 IEEE International Conference on Computer Vision
  Workshops (ICCVW)}, pages 270--279, 2017.

\bibitem{OpenLogo2018}
Hang Su, Xiatian Zhu, and Shaogang Gong.
\newblock Open logo detection challenge.
\newblock In {\em British Machine Vision Conference}, 2018.

\bibitem{Teh2020}
Eu~Wern Teh, Terrance DeVries, and Graham~W Taylor.
\newblock Proxynca++: Revisiting and revitalizing proxy neighborhood component
  analysis.
\newblock {\em arXiv preprint arXiv:2004.01113}, 2020.

\bibitem{tursun2015metu}
Osman Tursun and Sinan Kalkan.
\newblock Metu dataset: A big dataset for benchmarking trademark retrieval.
\newblock In {\em 2015 14th IAPR International Conference on Machine Vision
  Applications (MVA)}, pages 514--517. IEEE, 2015.

\bibitem{tuzko2017open}
Andras T{\"u}zk{\"o}, Christian Herrmann, Daniel Manger, and J{\"u}rgen
  Beyerer.
\newblock Open set logo detection and retrieval.
\newblock {\em arXiv preprint arXiv:1710.10891}, 2017.

\bibitem{LiTW2018}
Andras Tüzkö., Christian Herrmann., Daniel Manger., and Jürgen Beyerer.
\newblock Open set logo detection and retrieval.
\newblock In {\em Proceedings of the 13th International Joint Conference on
  Computer Vision, Imaging and Computer Graphics Theory and Applications -
  Volume 5: VISAPP,}, pages 284--292. INSTICC, SciTePress, 2018.

\bibitem{LogoDet3K2022}
Jing Wang, Weiqing Min, Sujuan Hou, Shengnan Ma, Yuanjie Zheng, and Shuqiang
  Jiang.
\newblock Logodet-3k: A large-scale image dataset for logo detection.
\newblock {\em ACM Trans. Multimedia Comput. Commun. Appl.}, 18(1), jan 2022.

\bibitem{yang2021tap}
Zhengyuan Yang, Yijuan Lu, Jianfeng Wang, Xi Yin, Dinei Florencio, Lijuan Wang,
  Cha Zhang, Lei Zhang, and Jiebo Luo.
\newblock Tap: Text-aware pre-training for text-vqa and text-caption.
\newblock In {\em Proceedings of the IEEE/CVF Conference on Computer Vision and
  Pattern Recognition}, pages 8751--8761, 2021.

\bibitem{zhai2019classification}
Andrew Zhai and Hao-Yu Wu.
\newblock Classification is a strong baseline for deep metric learning, 2019.

\bibitem{zhu2020simple}
Qi Zhu, Chenyu Gao, Peng Wang, and Qi Wu.
\newblock Simple is not easy: A simple strong baseline for textvqa and
  textcaps.
\newblock {\em arXiv preprint arXiv:2012.05153}, 2, 2020.

\end{thebibliography}
}

\newpage
\appendix

\section{Class Agnostic Logo Detector Model}
\begin{figure}
  \centering
  \includegraphics[width=8cm]{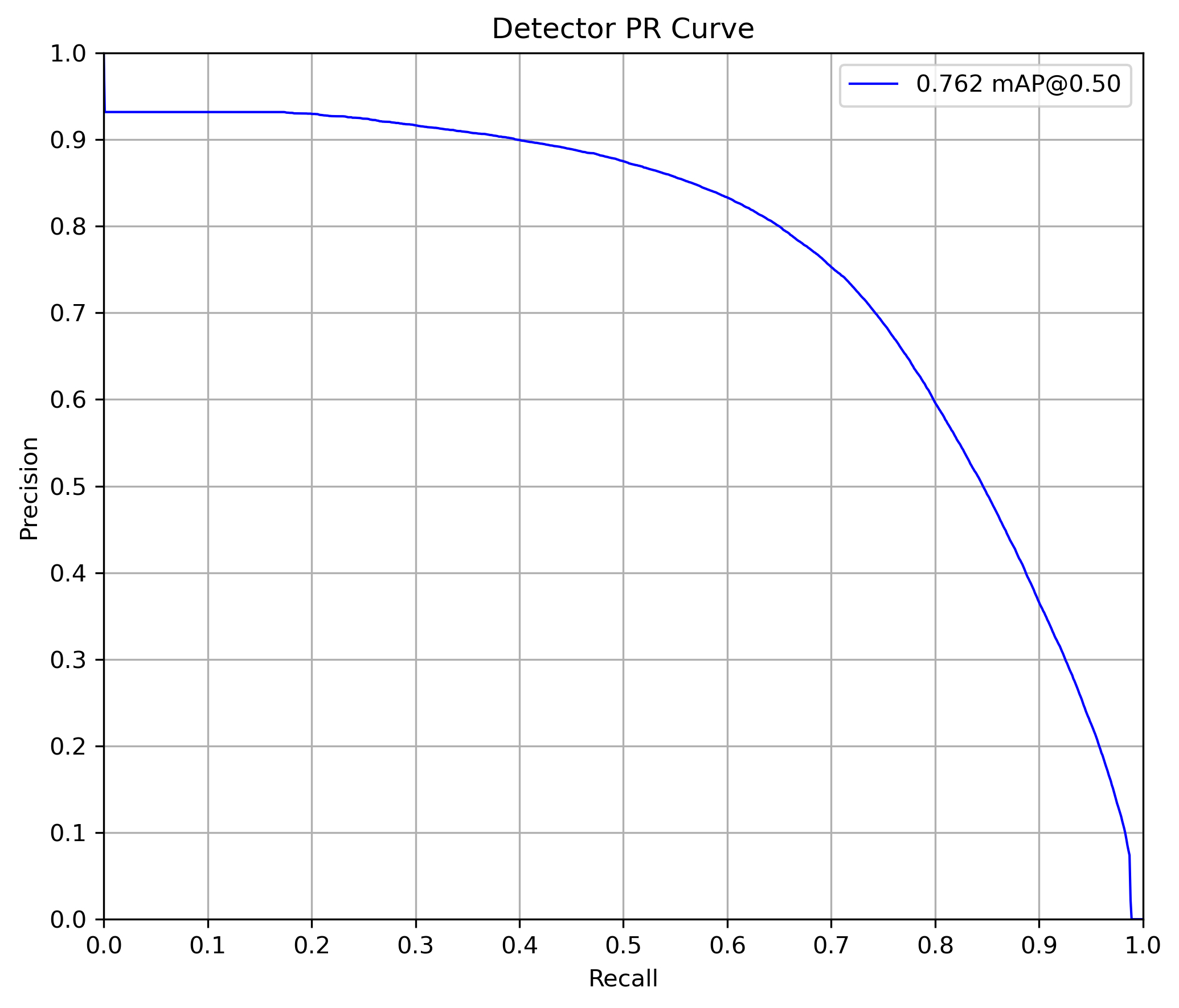}
  \caption{YoloV4 Logo Detector PR Curve}
  \label{fig:detectorpr}
\end{figure}

In conjunction with the logo embedding model presented in this work, a class agnostic logo detection model allows one to make end-to-end predictions on an input image.
We train a YoloV4 model on the PL2K dataset, consisting of $185247$, $46312$, and $57970$ Amazon product images in the train, validation, and test splits, respectively.
Each image is annotated with axis-aligned bounding boxes to localize a logo region.
We used an input size of $512$ pixels and the medium depth version of the architecture, containing $23.4$ million parameters.
Average precision at $0.5$ IoU is $0.762$ and its corresponding precision-recall curve on the test split is shown in Figure \ref{fig:detectorpr}.

\begin{figure*}
  \centering
  \begin{tabular}{c@{\hskip 0.1in}c@{\hskip 0.1in}c}
    \includegraphics[width=4cm]{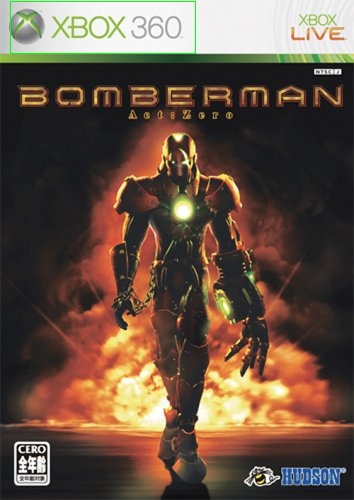} & \includegraphics[width=4cm]{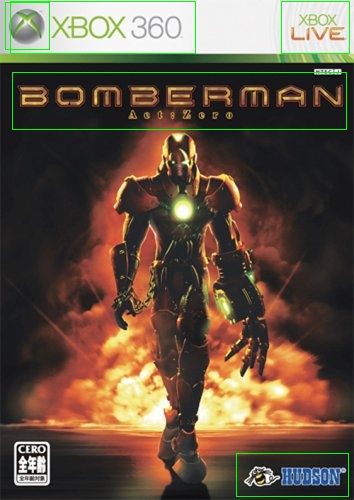} & \includegraphics[width=4cm]{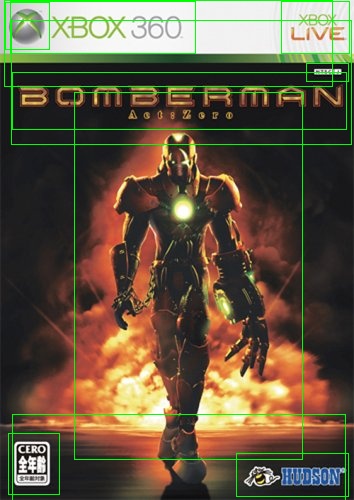} \\
    \textsf{0.4} & \textsf{0.1} & \textsf{0.01} \\
  \end{tabular}
  \caption{Logo detector predicted bounding boxes on an input image with different applied objectness score thresholds as specified below each image}
  \label{fig:detectoroutputs}
  \end{figure*}

In Figure \ref{fig:detectoroutputs} we also show several predictions on the same image when varying the objectness score threshold.
When using a logo embedding model together with the detector in order to identify logos in an input image, we observe that operating the detector at a lower threshold can improve overall recall of the system.
However, more extraneous regions are surfaced at such a lower threshold, which places greater importance on the ability of the embedding model to distinguish hard-negative background regions from actual logos.

\section{Cleaning of OpenLogoDet3K47 Dataset}
After merging the various public logo datasets of which OpenLogoDet3K47 consists, we cleaned the resulting union via the following steps:
\begin{enumerate}
\item Force all class names to lower case, replace ``-" or `` " with ``\_" and ``'" with an empty string.
\item Merge some obvious classes such as $\textsf{\small lv} \mapsto \textsf{\small louisvuitton}$, $\textsf{\small cocacola} \mapsto \textsf{\small coca\_cola}$, $\textsf{\small northface} \mapsto \textsf{\small the\_north\_face}$
\item For certain classes where there are text/symbol child classes from one source and not in another, we carefully reassign labels or merge them to minimize overall label noise.
\item Remove duplicate images within the same class.
\end{enumerate}
After this process, we have $3276$ classes with $188244$ images and $235738$ objects.
For our experiments, we filter out any image regions with a minimum side length less than $10$ pixels and we remove classes with fewer than $20$ instances.
This yields $2714$ classes with $181552$ images and $227176$ objects.

\section{Fixed Hyperparameters}
Table \ref{table:hyperparameters} shows hyperparameters that were fixed depending on model architecture for the experiments performed in the main paper.
\begin{table}
  \begin{center}
  \caption{Fixed hyperparameters}
  \label{table:hyperparameters}
  \small
  \begin{tabular}{lccc}
  \hline
  \noalign{\smallskip}
   & ViT & ResNet50 & CLIP RN50 \\
   \noalign{\smallskip}
   \hline
  Temperature Scaling $\sigma$ & $0.06$ & $0.06$ & $0.06$ \\
  Trunk Weight Decay & $0.2$ & $0.2$ & $0.2$  \\
  Last FC Weight Decay & $0.001$ & $0.001$ & $0.001$ \\
  Proxy Weight Decay & 0 & 0 & 0\\
  Adam $\beta_{1, \{\mathrm{trunk}, \mathrm{fc}, \mathrm{proxy}\}}$ & $0.9$ & $0.9$ & $0.9$ \\
  Adam $\beta_{2, \{\mathrm{trunk}, \mathrm{fc}\}}$ & $0.98$ & $0.999$ & $0.999$ \\
  Adam $\beta_{2, \mathrm{proxy}}$ & $0.999$ & $0.999$ & $0.999$ \\
  Adam $\epsilon_{\{\mathrm{trunk}, \mathrm{fc}\}}$ & $10^{-6}$ & $10^{-8}$ & $10^{-8}$ \\
  Adam $\epsilon_{\mathrm{proxy}}$ & $1$ & $1$ & $1$ \\
  \hline
  \end{tabular}
  \end{center}
\end{table}

\section{Additional ViT Comparison On OpenLogoDet3K47}
The test split of OpenLogoDet3K47 contains a total of $48098$ images.
Table \ref{table:vitmistakecomparison} shows a summary of mistakes by the best ViT embedder pre-trained on ImageNet and the best ViT embedder pre-trained on image-text data.
In particular, we see that the image-text pre-trained ViT model has over $3$ times as many correct predictions when the ImageNet pre-trained ViT was incorrect as vice versa.
\begin{table*}
\begin{center}
\caption{Counts of correct and incorrect predictions (using the closest neighbor) by two ViT embedding models. $|TP \cap FP_{\mathrm{other}}|$ indicates the size of the set of true positives from the given model intersected with the set of false positives from the other model. $FP_{1} \cap FP_{2}$ indicates the set of query images for which both models predicted incorrectly}
\label{table:vitmistakecomparison}
\small
\begin{tabular}{ccccccc}
Model & Pre-Training & $|TP|$      & $|FP|$        & $|TP \cap FP_{\mathrm{other}}|$ & $|FP_{1} \cap FP_{2}|$ & Precision\\
\hline
ViT   & OpenAI IT    & 47461   & 637       & \bf{826}                      & 352            & \bf{0.9867} \\ 
ViT   & ImageNet     & 46920   & 1178      & 285                           & 352            & 0.9755 \\
\end{tabular}
\end{center}
\end{table*}

\begin{table*}
  \begin{center}
  \caption{Recall@1 performance for best ViT model pre-trained on image-text data vs best ViT model pre-trained on ImageNet data. Both models trained on LogoDet3K train and val splits with test set held out.}
  \label{table:logodet3kresults2}
  \small
  \begin{tabular}{ccccccccccccc}
  Model & Pre-Training & \multicolumn{2}{c}{LogoDet3K Test} & \multicolumn{3}{c}{OpenLogo} & \multicolumn{2}{c}{BelgaLogo} & \multicolumn{2}{c}{FlickrLogos-47} & \multicolumn{2}{c}{LiTW}\\
        &              & QvG & All & QvG & All & Text & All & Text & All & Text & All & Text\\
  \hline
  ViT   & OpenAI IT    & \bf{0.9836} & \bf{0.9886} & \bf{0.9371} & 0.9629 & \bf{0.9568} & 0.9797 & \bf{0.9784}  & 0.9834 & \bf{0.9778} & 0.9391 & \bf{0.9456}\\
  ViT   & ImageNet     & 0.9622 & 0.9740 & 0.9305 & \bf{0.9675} & 0.9463 & \bf{0.9809} & 0.9753  & \bf{0.9879} & 0.9759 & \bf{0.9394} & 0.9169\\
  \end{tabular}
  \end{center}
\end{table*}
In Table \ref{table:logodet3kresults2} we compare performance of the image-text and ImageNet pre-trained ViT embedding models on several public logo datasets.
Each model was trained on the train and validation splits of the LogoDet3K datasets and evaluated on the test split.
This is the open-set regime where test classes in LogoDet3K are unseen.
On the most sizeable of these public datasets, OpenLogo and LogosInTheWild, we see $1\%$ and $3\%$ increases in recall@1 performance when restricted to classes containing ``text" in the name.
We further note that recall@1 performance on LogoDet3K is over $2\%$ better in the query versus gallery setting.
In the few cases where the ImageNet pre-trained model performs better in Table \ref{table:logodet3kresults2}, the difference is relatively small.

\begin{figure*}
  \centering
  \begin{tabular}{cccccc}
    \textsf{Query} & \textsf{Correct} & \textsf{Incorrect} & \textsf{Query} & \textsf{Correct} & \textsf{Incorrect}\\
    \noalign{\smallskip}
    \includegraphics[width=1.5cm,height=1.5cm]{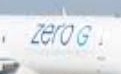} & \includegraphics[width=1.5cm,height=1.5cm]{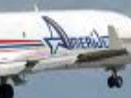} & \includegraphics[width=1.5cm,height=1.5cm]{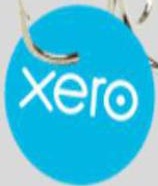} & \includegraphics[width=1.5cm,height=1.5cm]{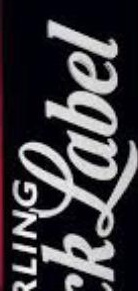} & \includegraphics[width=1.5cm,height=1.5cm]{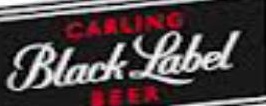} & \includegraphics[width=1.5cm,height=1.5cm]{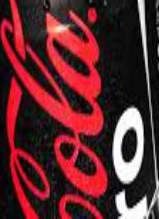}\\
    \tiny{\textsf{amerijet\_international}} &  & \tiny{\textsf{xero}} & \tiny{\textsf{carling\_black\_label}} & & \tiny{\textsf{coca\_cola\_zero}}\\
    \noalign{\smallskip}
    \includegraphics[width=1.5cm,height=1.5cm]{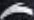} & \includegraphics[width=1.5cm,height=1.5cm]{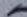} & \includegraphics[width=1.5cm,height=1.5cm]{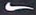} & \includegraphics[width=1.5cm,height=1.5cm]{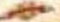} & \includegraphics[width=1.5cm,height=1.5cm]{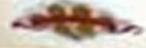} & \includegraphics[width=1.5cm,height=1.5cm]{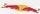}\\
    \tiny{\textsf{anta}} &  & \tiny{\textsf{nike}} & \tiny{\textsf{chateau\_real}} & & \tiny{\textsf{redbull}}\\
    \noalign{\smallskip}
    \includegraphics[width=1.5cm,height=1.5cm]{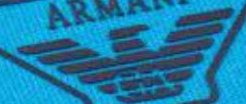} & \includegraphics[width=1.5cm,height=1.5cm]{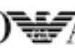} & \includegraphics[width=1.5cm,height=1.5cm]{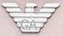} &     \includegraphics[width=1.5cm,height=1.5cm]{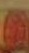} & \includegraphics[width=1.5cm,height=1.5cm]{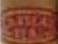} & \includegraphics[width=1.5cm,height=1.5cm]{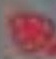} \\
    \tiny{\textsf{armani\_junior}} &  & \tiny{\textsf{armani}} & \tiny{\textsf{chateau\_real}} & & \tiny{\textsf{ariat}}\\
    \noalign{\smallskip}
    \includegraphics[width=1.5cm,height=1.5cm]{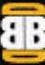} & \includegraphics[width=1.5cm,height=1.5cm]{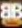} & \includegraphics[width=1.5cm,height=1.5cm]{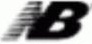} & \includegraphics[width=1.5cm,height=1.5cm]{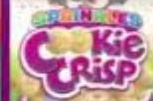} & \includegraphics[width=1.5cm,height=1.5cm]{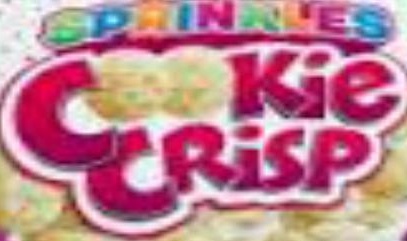} & \includegraphics[width=1.5cm,height=1.5cm]{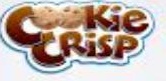}\\
    \tiny{\textsf{back\_yard\_burgers}} &  & \tiny{\textsf{new\_balance}} & \tiny{\textsf{cookie\_crisp\_sprinkles}} & & \tiny{\textsf{cookie\_crisp}}\\
    \noalign{\smallskip}
    \includegraphics[width=1.5cm,height=1.5cm]{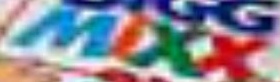} & \includegraphics[width=1.5cm,height=1.5cm]{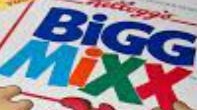} & \includegraphics[width=1.5cm,height=1.5cm]{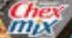} & \includegraphics[width=1.5cm,height=1.5cm]{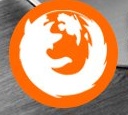} & \includegraphics[width=1.5cm,height=1.5cm]{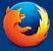} & \includegraphics[width=1.5cm,height=1.5cm]{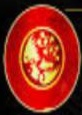}\\
    \tiny{\textsf{bigg\_mixx}} &  & \tiny{\textsf{chex\_mix}} & \tiny{\textsf{firefox}} & & \tiny{\textsf{xifeng}}\\
    \noalign{\smallskip}
    \includegraphics[width=1.5cm,height=1.5cm]{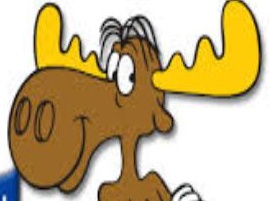} & \includegraphics[width=1.5cm,height=1.5cm]{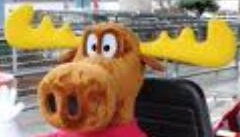} & \includegraphics[width=1.5cm,height=1.5cm]{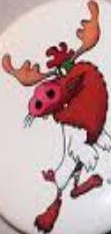} &     \includegraphics[width=1.5cm,height=1.5cm]{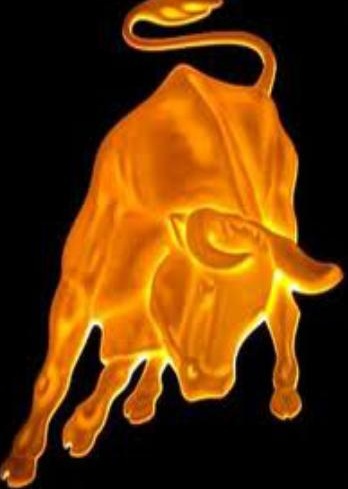} & \includegraphics[width=1.5cm,height=1.5cm]{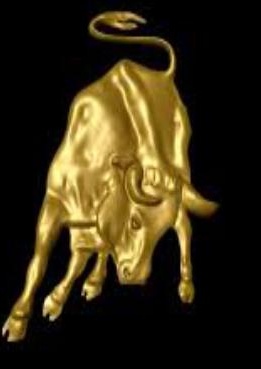} & \includegraphics[width=1.5cm,height=1.5cm]{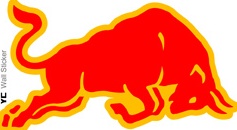}\\
    \tiny{\textsf{bullwinkles\_restaurant}} &  & \tiny{\textsf{bigg\_mixx}} & \tiny{\textsf{lamborghini}} & & \tiny{\textsf{redbull}}\\
    \noalign{\smallskip}
    \includegraphics[width=1.5cm,height=1.5cm]{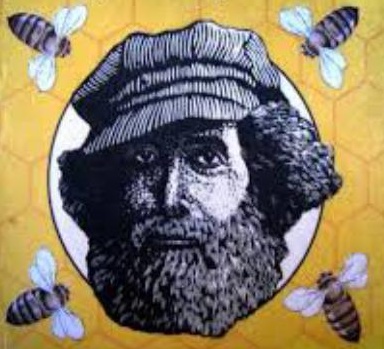} & \includegraphics[width=1.5cm,height=1.5cm]{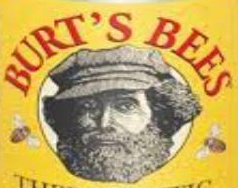} & \includegraphics[width=1.5cm,height=1.5cm]{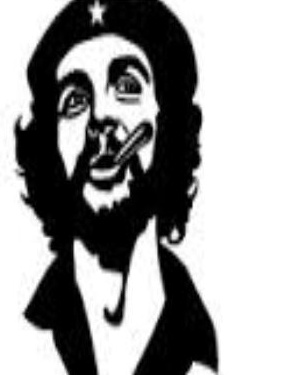} & \includegraphics[width=1.5cm,height=1.5cm]{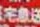} & \includegraphics[width=1.5cm,height=1.5cm]{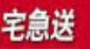} & \includegraphics[width=1.5cm,height=1.5cm]{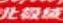}\\
    \tiny{\textsf{burts\_bees}} &  & \tiny{\textsf{che}} & \tiny{\textsf{zjs\_express}} & & \tiny{\textsf{beijirong}}\\
  \end{tabular}
  \caption{Sample images from the OpenLogoDet3K47 dataset test split where the image-text pre-trained ViT logo embedder predicts incorrectly while the ImageNet pre-trained ViT predicts correctly. Columns 1+4: query image, columns 2+5: ImageNet-pre-trained ViT model's correct logo retrieval, columns 3+6: Image-text pre-trained ViT model's incorrect logo retrieval}
  \label{fig:timmvitvsopenaivitvisual}
  \end{figure*}
  We noticed among the entire set of $285$ query images from the test split where the image-text pre-trained model made an incorrect prediction and the ImageNet1K pre-trained model was correct, many such images were very small in size or blurry to the point of ambiguity.
  We also noticed some query images that seem to be assigned to the wrong class, such as the top left image in Figure \ref{fig:timmvitvsopenaivitvisual}.
  Moreover, there is some label redundancy that lead to mistakes, such as the three \textsf{armani}-related classes: \textsf{armani}, \textsf{armani\_junior}, and \textsf{armani\_exchange}.
  Finally, in many cases when the image-text pre-trained model was incorrect, it was still able to match to a logo with similar letters and/or text style.
  Several of these comparison images are shown in Figure \ref{fig:timmvitvsopenaivitvisual}.\

\end{document}